\documentclass[10pt,twocolumn,letterpaper]{article}

\usepackage{cvpr}
\usepackage{times}
\usepackage{epsfig}
\usepackage{graphicx}
\usepackage{amsmath}
\usepackage{amssymb}

\usepackage[T1]{fontenc}    
\usepackage{url}            
\usepackage{booktabs}       
\usepackage{amsfonts}       
\usepackage{nicefrac}       
\usepackage{microtype}      
\usepackage{enumitem}

\usepackage{subcaption}
\usepackage{multirow}
\usepackage{comment}

\usepackage{algorithm}
\usepackage{algorithmic}
\usepackage{array}
\usepackage{bm}

\newcommand{\Tref}[1]{Table~\ref{#1}}

\newcommand{\Fref}[1]{Fig.~\ref{#1}}
\newcommand{\Sref}[1]{Sec.~\ref{#1}}

\newcommand{\datasetclipart}{Clipart1k}
\newcommand{\datasetwatercolor}{Watercolor2k}
\newcommand{\datasetcomic}{Comic2k}
\newcommand{\bam}{BAM!}
\newcommand{\lvl}{~~~}
\graphicspath{{./img/}}

\usepackage[pagebackref=true,breaklinks=true,letterpaper=true,colorlinks,bookmarks=false]{hyperref}

\cvprfinalcopy 

\def\cvprPaperID{1164} 

\begin{document}

\title{Cross-Domain Weakly-Supervised Object Detection \\ through Progressive Domain Adaptation}

\author{
	Naoto Inoue~~~~~Ryosuke Furuta~~~~~Toshihiko Yamasaki~~~~~Kiyoharu Aizawa\\
	The University of Tokyo, Japan\\
	{\tt\small \{inoue, furuta, yamasaki, aizawa\}@hal.t.u-tokyo.ac.jp}
}

\maketitle

\begin{abstract}
Can we detect common objects in a variety of image domains without instance-level annotations?
In this paper, we present a framework for a novel task, cross-domain weakly supervised object detection, which addresses this question.
For this paper, we have access to images with instance-level annotations in a source domain (e.g., natural image) and images with image-level annotations in a target domain (e.g., watercolor).
In addition, the classes to be detected in the target domain are all or a subset of those in the source domain.
Starting from a fully supervised object detector, which is pre-trained on the source domain, we propose a two-step progressive domain adaptation technique by fine-tuning the detector on two types of artificially and automatically generated samples.
We test our methods on our newly collected datasets\footnote{Datasets and codes are available at \url{https://naoto0804.github.io/cross_domain_detection}} containing three image domains, and achieve an improvement of approximately 5 to 20 percentage points in terms of mean average precision (mAP) compared to the best-performing baselines.
\end{abstract}

\section{Introduction}
Object detection is a task to localize instances of particular object classes in an image.
It is a fundamental task and has advanced rapidly due to the development of convolutional neural networks (CNNs).
Best-performing detectors~\cite{girshick2015fast,ren2015faster,liu2016ssd,redmon2016yolo9000,li2016r,lin2017focal} are fully supervised detectors (FSDs).
They are highly data-hungry and typically learned from many images with instance-level annotations.
An instance-level annotation is composed of a label (i.e., the object class of an instance) and a bounding box (i.e., the location of the instance).

While object detection in a natural image domain has achieved outstanding performance, less attention has been paid to the detection in other domains such as watercolor.
This is because it is often difficult and unrealistic to construct a large dataset with instance-level annotations in many image domains.
There are many obstacles such as lack of image sources, copyright issues, and the cost of annotation.

We tackle a novel task, cross-domain weakly supervised object detection.
The task is described as follows: (i) \textit{instance}-level annotations are available in a \textit{source} domain; (ii) only \textit{image}-level annotations are available in a \textit{target} domain; (iii) the classes to be detected in the target domain are all or a subset of those in the source domain.
The objective of the task is to detect objects as accurately as possible in the target domain under these conditions by using sufficient instance-level annotations in the source domain and a small number of image-level annotations in the target domain.
This assumption is reasonable as it is easier to collect image-level annotations than instance-level annotations from existing datasets or an image search engine.

\begin{figure}[t]
	\centering
    \includegraphics[width=\hsize]{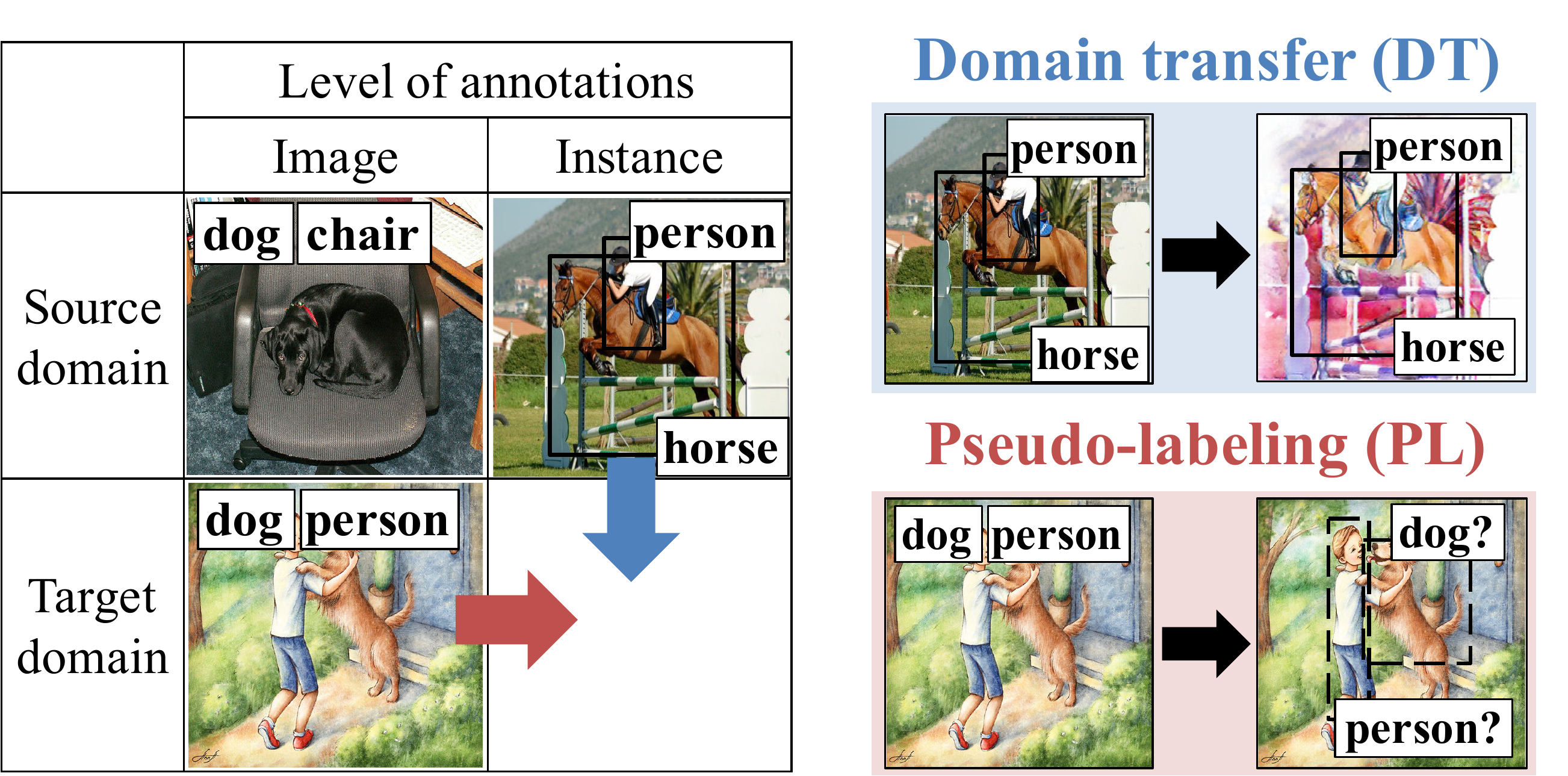}
    \caption{Left: the situation of the cross-domain weakly supervised object detection; Right: Our methods to generate instance-level annotated samples in the target domain.}
	\label{fig:teaser}
\end{figure}

We will describe a framework to solve the proposed task.
Starting from an FSD trained on images with instance-level annotations in the source domain, we fine-tune the FSD in the target domain, as this is the most straightforward and promising approach.
However, there are no \textit{instance}-level annotations available in the target domain.
Instead, as shown in \Fref{fig:teaser}, we present two methods to generate images with \textit{instance}-level annotations artificially and automatically, and fine-tune the FSD on them.
The first method, \textit{domain transfer} (DT), is used to generate images that look like those in the target domain from images in the source domain having \textit{instance}-level annotations.
This generation is achieved by image-to-image translation methods from unpaired examples such as CycleGAN~\cite{zhu2017unpaired}.
The second method, \textit{pseudo-labeling} (PL), is used to generate pseudo \textit{instance}-level annotations.
Given images with \textit{image}-level annotations in the target domain and the FSD which is fine-tuned on the artificially generated samples by DT, these annotations and predictions of the FSD are combined.
We achieve a two-step progressive domain adaptation by sequentially fine-tuning the FSD on the artificially generated samples.
Our framework is general to cross-domain weakly supervised object detection across any image domain and is relatively scalable to many classes and instances.

Since there is no dataset for the target domain that is suitable to evaluate the proposed task, we construct new datasets with instance-level annotations, which we call ~\textit{\datasetclipart},~\textit{\datasetwatercolor}, and~\textit{\datasetcomic}.
Each dataset comprises 1,000, 2,000, and 2,000 images of clipart, watercolor, and comic, respectively.
The validity of our methods is demonstrated using these datasets.
We show that the proposed two-step adaptation achieves an improvement of approximately 5 to 20 percentage points as compared to the best-performing baselines' mAP across all datasets.
We believe that this paper itself can be a strong baseline for cross-domain weakly supervised object detection.

Our main contributions are as follows:
\begin{itemize}[wide=0pt]
\setlength{\itemsep}{0cm}
    \item We propose a framework for a novel task, cross-domain weakly supervised object detection.
We achieve a two-step progressive domain adaptation by sequentially fine-tuning the FSD on the artificially generated samples by the proposed domain transfer and pseudo-labeling.
    \item We construct novel, fully instance-level annotated datasets with multiple instances of various object classes across three domains that are far from natural images.
    \item Our experimental results show that our framework outperforms the best-performing baselines by approximately 5 to 20 percentage points in terms of mAP.
\end{itemize}

\begin{figure*}[t]
	\begin{minipage}[t]{0.36\hsize}
		\centering
    	\includegraphics[height=6cm]{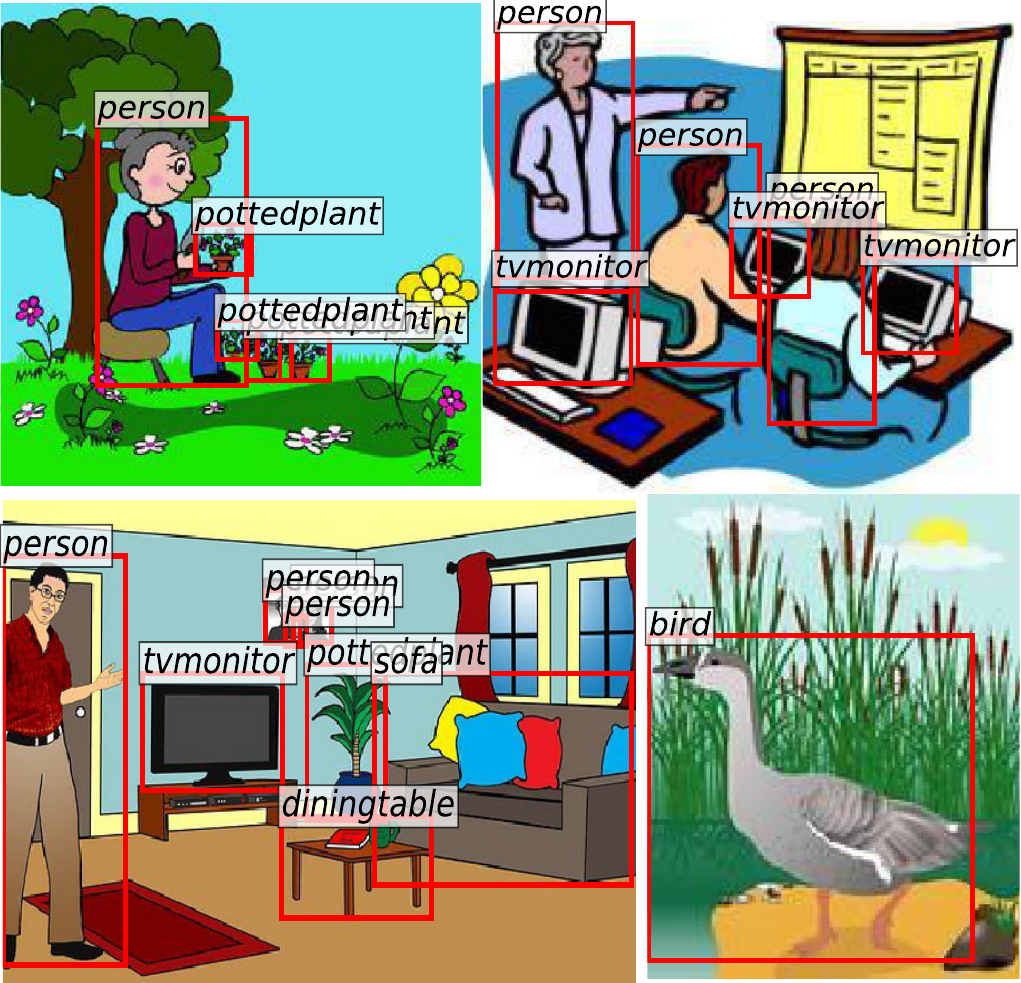}
    	\vspace{-0.35cm}
		\subcaption{\datasetclipart}
		\label{fig:gt_clipart}
	\end{minipage}
	\hfill
	\begin{minipage}[t]{0.30\hsize}
		\centering
    	\includegraphics[height=6.5cm]{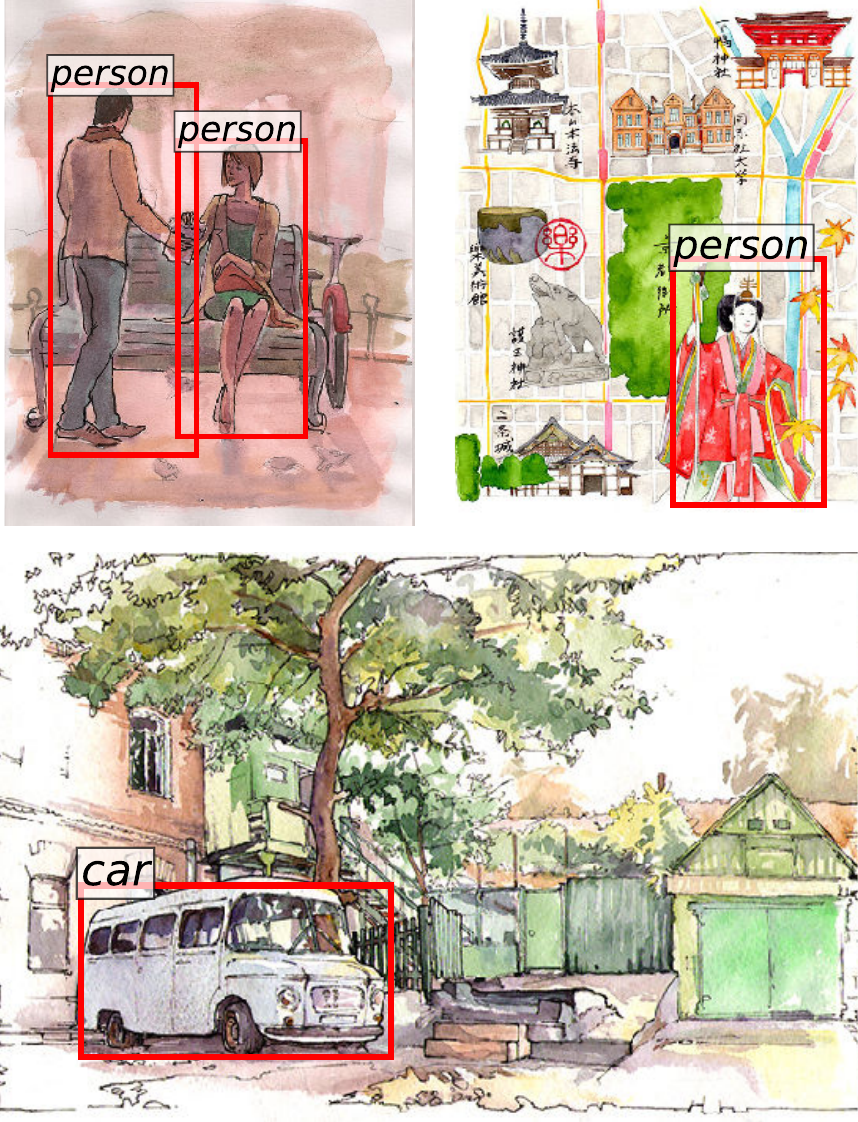}
    	\vspace{0.05cm}
		\subcaption{\datasetwatercolor}
		\label{fig:gt_watercolor}
	\end{minipage}
	\hfil
	\begin{minipage}[t]{0.3\hsize}
		\centering
    	\includegraphics[height=6.5cm]{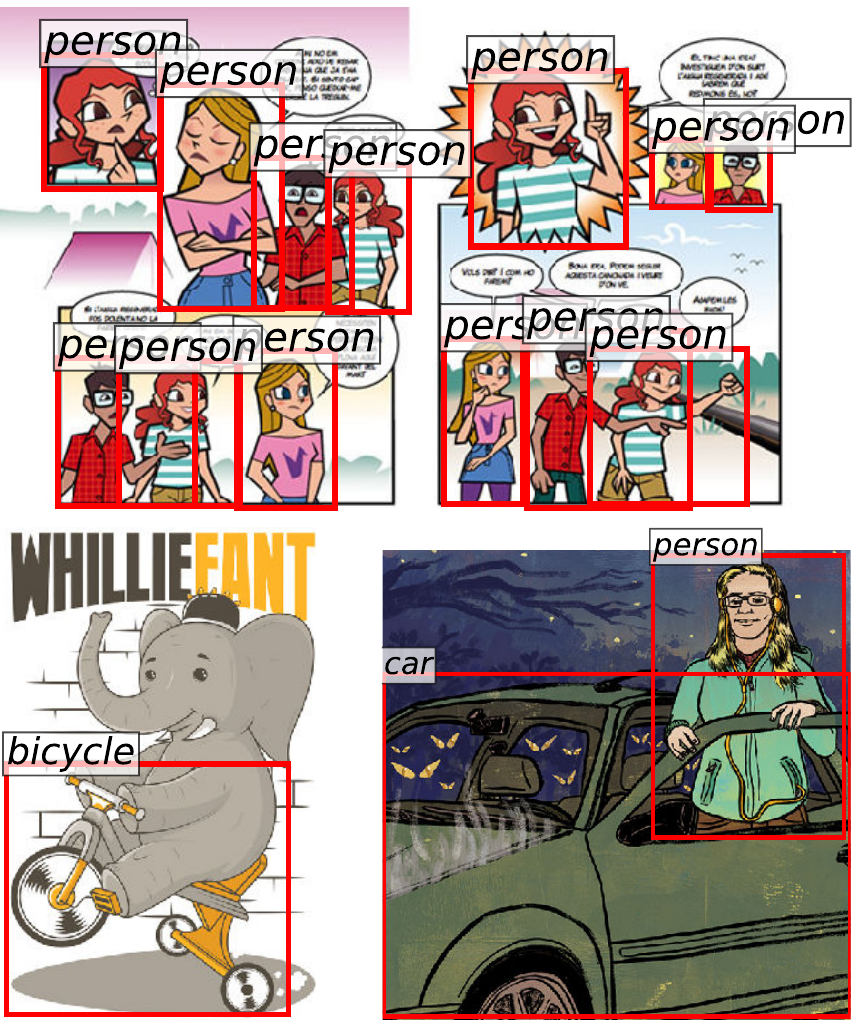}
    	\vspace{-0.4cm}
		\subcaption{\datasetcomic}
		\label{fig:gt_comic}
	\end{minipage}
    \caption{Examples of datasets that we collected across three domains; The images usually contain not only the target objects but also various other objects and complex backgrounds.}
    \label{fig:dataset_images}
\end{figure*}
\section{Related Work}
\subsection{Fully Supervised Detection}
\label{subsec:fully_supervised_detection}
Standard methods in fully supervised object detection, such as R-CNN~\cite{girshick2014rich}, Fast R-CNN~\cite{girshick2015fast}, and Faster R-CNN~\cite{ren2015faster}, are based on a two-stage approach: generating region proposals and then, classifying them.
Recently, single-stage object detectors such as SSD~\cite{liu2016ssd}, YOLOv2~\cite{redmon2016yolo9000}, and RetinaNet~\cite{lin2017focal} have also emerged.
All of these detectors require large datasets with instance-level annotations such as PASCAL VOC~\cite{everingham2010pascal}, Microsoft Common Objects in Context (MSCOCO)~\cite{lin2014microsoft}, and OpenImages~\cite{openimages}.

Dataset construction for a new image domain becomes harder as the number of images and classes increases.
\cite{su2012crowdsourcing} reported that it took 35 seconds for a worker to annotate one bounding box.
Recently,~\cite{papadopoulos2017extreme} reduced it to 7 seconds through extreme clicking, while it still takes much time and effort to obtain large-scale datasets.
On the contrary, our framework does not require instance-level annotations for the new target domain at all.

\subsection{Weakly Supervised Detection}
\label{subsec:weakly_supervised_detection}
One possible approach addressing the lack of large-scale instance-level annotations for object detection is to use a weakly supervised detector (WSD).
In weakly supervised object detection, only pairs of an image and an image-level annotation (i.e., labels of objects in each image) are provided for training.
Many existing methods are built upon region-of-interest (RoI) extraction methods such as selective search~\cite{uijlings2013selective}.
Feature extraction for each region, region selection, and classification of the selected region are performed through multiple instance learning (MIL)~\cite{song2014learning,gokberk2014multi,song2014weakly,bilen2015weakly,li2016weakly} or two-stream CNN~\cite{bilen2016weakly,kantorov2016contextlocnet,tang2017multiple}.
However, WSDs are poor at accurately localizing the object boundary.
Our framework uses image-level annotations in the target domain by pseudo-labeling the image.

\subsection{Cross-domain Object Detection}
Using an object detector that is neither trained nor fine-tuned for the target domain causes a significant drop in performance as shown in~\cite{wilber2017bam}.
Therefore, adapting the detector with the help of some information on the target domain is essential. 
\cite{hoffman2015detector,chen2015webly} are the some of the best works closely related to this paper.
Our methods and~\cite{hoffman2015detector} are similar as they propose to learn from a combination of instance- and image-level annotations.
However, we address the adaptation of the detector from one domain to another, whereas~\cite{hoffman2015detector} addresses the classifier-to-detector adaptation for weakly labeled object classes within one domain.
This paper and~\cite{chen2015webly} are similar as they tackle the adaptation of the detector from one domain to another.
However, only image-level annotations are available in the source domain in~\cite{chen2015webly}.
This is the first work to propose the cross-domain weakly supervised object detection.

For evaluating the cross-domain object detection method, the existing datasets for detecting common objects in various domains seem to have limitations.
People-Art~\cite{westlake2016detecting} is used only for single-class detection in an artwork.
Photo-Art~\cite{wu2014learning} assumes only one instance per image, which is unrealistic.
Besides, we introduce the fully instance-level annotated datasets for object detection which comprises multiple common classes to be detected various visual domains.

\subsection{Unsupervised Domain Adaptation}
Unsupervised domain adaptation (UDA) in an image is a task used for learning domain invariant models, where pairs of an image and annotation are available in the source domain while only images are available in the target domain.
Previous works for UDA in image classification is mostly distribution-matching-based, in which features extracted from two domains are made to closely resemble each other using the maximum mean discrepancy (MMD)~\cite{gretton2012kernel} or a domain classifier network~\cite{long2015learning,ganin2016domain,long2016unsupervised,tzeng2017adversarial}.
Although current distribution-matching-based methods are applicable, it is primarily challenging to fully align the distribution for tasks that require structured outputs, such as object detection.
This is because it is essential to keep the spatial information in the feature map.
Our framework employs image-to-image translation and fine-tuning to avoid this problem.

\section{Dataset}
\label{sec:dataset}
Our objective is to detect objects in a target domain by adapting an FSD that is originally trained on a source domain.
The classes to be detected in the target domain are all or a subset of the classes defined in the dataset which is in the source domain.
In this paper, PASCAL VOC~\cite{everingham2010pascal}, which contains twenty classes, was used for the source domain, natural image.
As no suitable dataset for the target domain of our task was available, we constructed three original datasets, \datasetclipart, \datasetwatercolor, and~\datasetcomic~using Amazon Mechanical Turk.

\begin{table}[t]
  \caption{List of the datasets that we constructed for the target domains in this paper.}
  \label{tbl:comparison_of_datasets}
  \centering
  \begin{tabular}{@{}lrrr@{}} \toprule
    Dataset & \#classes & \#images & \#instances \\ \midrule
    \datasetclipart & 20 & 1,000 & 3,165 \\
    \datasetwatercolor & 6 & 2,000 & 3,315 \\
    \datasetcomic & 6 & 2,000 & 6,389 \\ \bottomrule
  \end{tabular}
\end{table}
Examples of the images are shown in \Fref{fig:dataset_images}. 
These images usually contain multiple objects per image, and
some instances are small or partially occluded by the other objects.
The statistics of the three datasets are shown in \Tref{tbl:comparison_of_datasets}.
We collected a total of 5,000 images and 12,869 instance-level annotations.
We believe these datasets are good benchmarks not only for domain adaptation but also for fully and weakly supervised, semi-supervised detection tasks.
For the more detailed statistics, please refer to the supplementary material.
In the following subsection, we will briefly describe each dataset, and the data collection method.

\subsection{\datasetclipart}
In~\datasetclipart, the target domain classes to be detected were the same as those in the source domain.
All the images for a clipart domain were collected from one dataset (i.e., CMPlaces~\cite{castrejon2016learning}) and two image search engines (i.e., Openclipart\footnote{https://openclipart.org/} and Pixabay\footnote{https://pixabay.com/}).
When we collected the images from the search engines, we used 
queries of the 205 scene classes (e.g., pasture) used in CMPlaces to collect various objects and scenes with complex backgrounds.

\subsection{\datasetcomic~and~\datasetwatercolor}
\label{subsec:dataset_comic_and_watercolor}
In~\datasetcomic~and~\datasetwatercolor, the classes to be detected in the target domain were the subset of those in the source domain.
The images were collected from~\bam~\cite{wilber2017bam}.
In~\bam~, millions of images with slightly noisy ($80 \%-90 \%$ in precision) image-level attributes regarding object classes, domain, and emotion are provided in a human-in-the-loop fashion.
Specifically, the target classes are \texttt{bicycle}, \texttt{bird}, \texttt{cat}, \texttt{car}, \texttt{dog}, and \texttt{person}, which are representative of the intersection of the classes in VOC and those in~\bam.
We chose the watercolor and comic domains as the other domains in~\bam~are not suitable for object detection.
For example, oil paint images are unsuitable as they usually depict a single person in the center of the image, making object detection a trivial task.

As collecting instance-level annotations for all images in~\bam~is difficult, we annotated the images in the following way: 
(i) images that contained at least one of the six target classes were extracted.
Note that we relied on the labels provided by~\bam~and did not conduct any other filtering process.
We obtained 17,814 and 52,790 images for watercolor and comic domains, respectively. 
(ii) as many as 2,000 images are randomly sampled and assigned instance-level annotations for each domain.
The remaining 15,814 and 50,790 images, which we called \texttt{extra} dataset, are still useful as they possess many image-level annotations.
Although they are noisy and incomplete, they provided room for further improvement with respect to detector performance as shown in \Sref{subsec:results_bam}.

\section{Proposed Method}
\begin{figure}[t]
    \centering
    \includegraphics[width=\hsize]{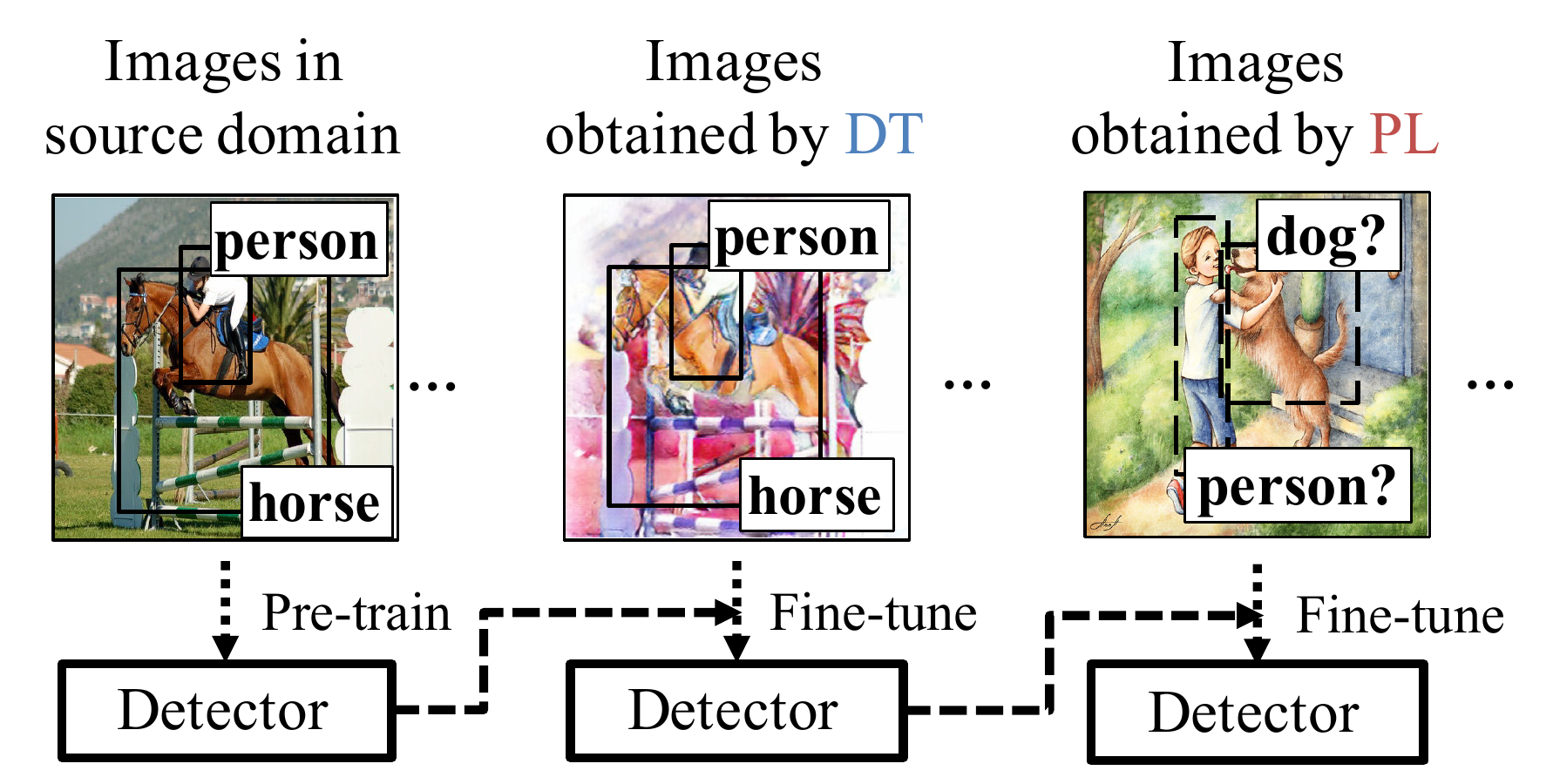}
    \caption{The workflow of our framework.}
    \label{fig:overview}
\end{figure}
We propose a framework to adapt an FSD that is pre-trained on a source domain.
The adaptation is achieved through fine-tuning the FSD on artificially generated samples with instance-level annotations in a target domain.
We propose two methods to generate the samples as shown in \Fref{fig:teaser}:
(i) domain transfer (DT), transferring images with instance-level annotations from the source domain to the target domain, and (ii) pseudo-labeling (PL), pseudo-labeling the images with image-level annotations in the target domain.
The samples generated by these two methods display different properties.
Although the samples generated by (i) are not high-quality images with respect to their similarities to target-domain images, bounding boxes are correctly annotated.
On the contrary, although the samples generated by (ii) do not have accurate bounding boxes, image quality is guaranteed as they are completely target-domain images.

We progressively fine-tune an FSD using these examples as shown in \Fref{fig:overview}.
First, we pre-train it while using instance-level annotations in the source domain.
Second, we fine-tune it while using the images obtained by DT.
Lastly, we fine-tune it while using the images obtained by PL.
We would like to emphasize that the sequential execution of the two fine-tuning steps is critical as the performance of PL highly depends on the used FSD.

\subsection{Domain Transfer (DT)}
\label{subsec:domain_transfer}
The differences between the source and target domains tackled in this paper mainly lie in their low-level features, such as color and texture.
We generate images that look like those in the target domain to capture such differences and then, make the FSD robust to such differences by fine-tuning the FSD on the generated images.

To achieve this goal, an unpaired image-to-image translation method called CycleGAN~\cite{zhu2017unpaired} is employed.
With CycleGAN, the goal is to learn the mapping functions between two image domains $\mathcal{X}$ and $\mathcal{Y}$ with unpaired examples.
In practice, a mapping $G: \mathcal{X} \rightarrow \mathcal{Y}$ and an inverse mapping $F: \mathcal{Y} \rightarrow \mathcal{X}$ are jointly learned using CNN.
We train CycleGAN to learn the mapping functions between the source domain, $\mathcal{X}_{s}$, and the target domain, $\mathcal{X}_{t}$.
Once the mapping functions are trained, we convert images in the source domain that are used in the pre-training and obtain domain-transferred images that accompany instance-level annotations.
Using these images, the FSD is fine-tuned.

\subsection{Pseudo-Labeling (PL)}
In the target domain, if we use an FSD that is trained only on the source domain for object detection, then the FSD mainly fails because of confusion with other classes and backgrounds rather than inaccurate localization.
We will later show this trend in \Fref{fig:visualization_of_performance}.
Fine-tuning FSD on images obtained by PL dramatically reduces such confusion.
PL is simple and applicable in any FSD as it does not access intermediate layers of an FSD.

Formally, the objective of PL is to obtain a pseudo instance-level annotation $\bm{G}$ for each image $\bm{x}$ from the target domain $\mathcal{X}_{t}$.
Let $\bm{x} \in \mathbb{R}^{H \times W \times 3}$ denote an RGB image, where $H$ and $W$ are the image's height and width, respectively.
$\mathcal{C}$ indicates a set of object classes.
$\bm{z}$ indicates an image-level annotation: the set of classes in $\bm{x}$.
Further, $\bm{G}$ comprises $g=(\bm{b}, c)$, where $\bm{b} \in \mathbb{R}^{4}$ is a bounding box, and $c \in \mathcal{C}$.
First, we obtain FSD outputs $\bm{D}$.
$\bm{D}$ comprises each detection $d~=~(p, \bm{b}, c)$, where $c \in \mathcal{C}$ and $p \in \mathbb{R}$ indicates the probability of $\bm{b}$ belonging to $c$.
Second, for each class $c \in \bm{z}$, we take the top-1 confident detection $d~=~(p, \bm{b}, c) \in \bm{D}$ and add $(\bm{b}, c)$ to $\bm{G}$.
We fine-tune the FSD using pairs of $\bm{x}$ and $\bm{G}$.
Note that no layers of the FSD are replaced to preserve the original network's detection ability.
The FSD that is trained on images obtained by DT was subsequently fine-tuned on these images.

\section{Experiments}
\label{sec:experiment}
In \Sref{subsec:impl_metrics}, we explain the details of the implementations, the compared methods, and the evaluation metrics.
In \Sref{subsec:results_clipart}, we test our methods using~\datasetclipart~and conduct error analysis and ablation studies on the FSDs.
In \Sref{subsec:results_bam}, we confirm that our framework is generalized for a variety of domains using~\datasetwatercolor~and~\datasetcomic.
In \Sref{subsec:qualitative_results}, we show actual detection results and the generated domain-transferred images for further discussion.

\subsection{Implementation and Evaluation Metrics}
\label{subsec:impl_metrics}

Our methods were implemented using Chainer~\cite{tokui2015chainer}.
We evaluated our methods using average precision (AP) and its mean, i.e., mAP.

\paragraph{Dataset Arrangement for Training and Evaluation}
VOC2007-trainval and VOC2012-trainval~\cite{everingham2010pascal} were used as images in the source domain (i.e., natural image) in all the experiments.
For the target-domain images, the ones with instance-level annotations were used when discussing the performance gap between our methods and the ideal case quantitatively.
The target-domain images were split into training set and test set by a ratio of 1:1.
For the training set, the bounding box information was ignored to meet the proposed situation.
For the test set, the labels and the bounding boxes of the annotations were used to evaluate the performance of the compared methods and our methods.

\paragraph{Comparison}
We compared our methods against the following methods:
\begin{itemize}[wide=0pt]
\setlength{\itemsep}{0cm}
	\item Baseline: SSD300~\cite{liu2016ssd} was used as our baseline FSD.
We used the implementation provided by ChainerCV~\cite{ChainerCV2017}, and
we obtained an SSD300, which was pre-trained on VOC2007-trainval and VOC2012-trainval, and skipped the pre-training.
We followed the original paper on hyper-parameters of SSD300 unless specified.
The input images were resized to 300 $\times$ 300 in SSD300.
The IoU threshold for NMS (0.45) and the confidence threshold for discarding low confidence detections (0.01) were employed.
	\item Ideal case: 
In this case, we had access to the instance-level annotations for the training set of the target domain dataset.
We simply fine-tuned the baseline FSD using these annotations.
This experiment was to confirm the weak upper-bound performance of our methods.
	\item Weakly supervised detection (WSD): ContextLocNet (CLNet)~\cite{kantorov2016contextlocnet} and WSDDN~\cite{bilen2016weakly} were chosen as the compared WSDs. Each WSD was trained on the images with image-level annotations from the training set of the dataset in the target domain.
	\item Unsupervised domain adaptation (UDA): We tested one of the state-of-the-art UDA methods, ADDA~\cite{tzeng2017adversarial}.
We aligned the distributions at the \texttt{relu4\_3} layer in SSD300. We trained the model with a batch size of 32 and a learning rate of $1.0 \times 10^{-6}$ for 1,000 iterations using Adam~\cite{kingma2015adam}.
	\item Ensemble: In image classification, unweighted averaging, which uses the unweighted average of the output score / probability of all the base models as the output, is a reasonable approach to boost performance.
We accumulated all the detections produced by multiple detectors and applied non-maximum suppression (NMS) \footnote{Further details about NMS can be found in~\cite{felzenszwalb2010object}.} to them.
The parameters of NMS were the same as those used in SSD300.
\end{itemize}

\paragraph{Details of Training}
We trained CycleGAN with a learning rate of $1.0 \times 10^{-5}$ for the first ten epochs and a linear decaying rate to zero over the next ten epochs.
We followed the original paper on the other hyper-parameters of CycleGAN.
When fine-tuning SSD300, we employed a learning rate of $1.0 \times 10^{-5}$, which is the same as the final learning rate for the original SSD300 training.
Fine-tuning, using the images obtained by DT and PL, was conducted for one epoch and 10,000 iterations, respectively.

\begin{table*}[t]
    \caption{Comparison of all the methods in terms of AP [\%] using SSD300 as the baseline FSD in \datasetclipart. \texttt{Ensemble} denotes an ensemble of SSD300, CLNet, and WSDDN.}    
    \label{tbl:ap_on_each_class}
    {\renewcommand\arraystretch{1.1}
        \scalebox{0.88}[0.88]{
            \tabcolsep=1.5pt
            \centering
            \begin{tabular}{@{}lccccccccccccccccccccc@{}}\toprule
                & \multicolumn{20}{c}{AP for each class} & \\
                \cmidrule(r){2-21}        
                Method & aero & bike & bird & boat & bottle & bus & car & cat & chair & cow & table & dog & horse & mbike & person & plant & sheep & sofa & train & tv & mAP \\ \midrule        
                Baseline & 19.8 & 49.5 & 20.1 & 23.0 & 11.3 & 38.6 & 34.2 & 2.5 & 39.1 & 21.6 & 27.3 & 10.8 & 32.5 & 54.1 & 45.3 & 31.2 & 19.0 & 19.5 & 19.1 & 17.9 & 26.8 \\ \midrule
                
                \textit{Compared} & & & & & & & & & & & & & & & & & & & & & \\          
                \lvl WSDDN~\cite{bilen2016weakly} & 1.6 & 3.6 & 0.6 & 2.3 & 0.1 & 11.7 & 4.5 & 0.0 & 3.2 & 0.1 & 2.8 & 2.3 & 0.9 & 0.1 & 14.4 & 16.0 & 4.5 & 0.7 & 1.2 & 18.3 & 4.4 \\
                \lvl CLNet~\cite{kantorov2016contextlocnet} & 3.2 & 22.3 & 2.2 & 0.7 & 4.6 & 4.8 & 17.5 & 0.2 & 4.8 & 1.6 & 6.4 & 0.6 & 4.7 & 0.6 & 12.5 & 13.1 & 14.1 & 4.1 & 8.0 & 29.7 & 7.8\\
                \lvl Ensemble & 20.6 & 49.6 & 20.5 & 23.4 & 11.3 & 39.3 & 35.2 & 2.6 & 39.0 & 22.8 & 27.3 & 11.2 & 33.2 & 54.7 & 34.0 & 30.7 & 21.0 & 20.3 & 20.3 & 18.3  & 26.7 \\
                \lvl ADDA~\cite{tzeng2017adversarial} & 20.1 & 50.2 & 20.5 & 23.6 & 11.4 & 40.5 & 34.9 & 2.3 & 39.7 & 22.3 & 27.1 & 10.4 & 31.7 & 53.6 & 46.6 & 32.1 & 18.0 & 21.1 & 23.6 & 18.3 & 27.4 \\ \midrule
                
                \textit{Proposed} & & & & & & & & & & & & & & & & & & & & & \\          
                \lvl PL w/o label & 18.6 & 40.3 & 17.1 & 16.7 & 4.9 & 35.3 & 36.1 & 1.1 & 36.0 & 22.9 & 29.1 & 14.7 & 31.5 & 52.6 & 43.8 & 28.6 & 13.3 & 14.6 & 32.8 & 15.1 & 25.3\\       
                \lvl PL & 24.2 & 59.8 & 22.0 & 26.6 & 25.0 & 54.7 & \textbf{51.3} & 3.9 & 47.4 & 44.5 & 40.3 & 14.3 & 33.6 & 55.1 & 50.8 & 41.1 & \textbf{23.2} & 26.3 & 40.5 & 43.2 & 36.4\\   
                \lvl DT & 23.3 & 60.1 & 24.9 & 41.5 & 26.4 & 53.0 & 44.0 & \textbf{4.1} & 45.3 & 51.5 & 39.5 & 11.6 & 40.4 & 62.2 & 61.1 & 37.1 & 20.9 & \textbf{39.6} & 38.4 & 36.0 & 38.0\\
                \lvl DT+PL w/o label & 16.8 & 53.7 & 19.7 & 31.9 & 21.3 & 39.3 & 39.8 & 2.2 & 42.7 & 46.3 & 24.5 & 13.0 & 42.8 & 50.4 & 53.3 & 38.5 & 14.9 & 25.1 & 41.5 & 37.3 & 32.7 \\   
                \lvl DT+PL & \textbf{35.7} & \textbf{61.9} & \textbf{26.2} & \textbf{45.9} & \textbf{29.9} & \textbf{74.0} & 48.7 & 2.8 & \textbf{53.0} & \textbf{72.7} & \textbf{50.2} & \textbf{19.3} & \textbf{40.9} & \textbf{83.3} & \textbf{62.4} & \textbf{42.4} & 22.8 & 38.5 & \textbf{49.3} & \textbf{59.5} & \textbf{46.0} \\ \midrule    
                Ideal case & 50.5 & 60.3 & 40.1 & 55.9 & 34.8 & 79.7 & 61.9 & 13.5 & 56.2 & 76.1 & 57.7 & 36.8 & 63.5 & 92.3 & 76.2 & 49.8 & 40.2 & 28.1 & 60.3 & 74.4 & 55.4\\ \bottomrule          
            \end{tabular}   
    }}
\end{table*}
\subsection{Quantitative Results on \datasetclipart}
\label{subsec:results_clipart}
\Tref{tbl:ap_on_each_class} shows the comparison of AP for each class and mAP among our methods against the baseline FSD and the comparable methods.
We observe that SSD300 performs better than the WSDs in terms of mAP, although SSD300 is not trained to adapt to the target domain.
On the contrary, WSDs perform poorer due to insufficient data and their poor localization ability.
The ensembling of WSDs with SSD300 have almost no effect as shown in the case of \texttt{Ensemble}.
Conventional distribution-matching-based methods do not work well as shown in the case of \texttt{ADDA}.

The results of our methods based on SSD300 are shown in the bottom half of \Tref{tbl:ap_on_each_class}.
To quantify the relative contribution of each step, the performances of our methods are examined using different configurations.
\begin{itemize}[wide=0pt]
\setlength{\itemsep}{0cm}
	\item DT+PL: the proposed two-step fine-tuning.
	\item DT: only fine-tuning using images obtained by DT.
	\item PL: only fine-tuning using images obtained by PL. Note that the baseline FSD is used for PL.
\end{itemize}
PL provides an improvement of 9.6 percentage points improvement from the baseline SSD300 in terms of mAP.
Further, DT+PL achieves 46.0~\% in terms of mAP.
This result ensures that both of our methods work and are complementary.
The mAP of the combination of DT+PL is 19.2 percentage points higher than that of the baseline SSD300 and is approximately 18 percentage points greater than the ensemble of the detectors.
We emphasize that this performance is only 9.4 percentage points lower than \texttt{Ideal case}.

\paragraph{Ablation Study}
We considered a setting where we can obtain only images with no annotation in the target domain.

DT is applicable without any modification.
DT provides an improvement of 11.2 percentage points improvement from the baseline SSD300 in terms of mAP in \Tref{tbl:ap_on_each_class}.

PL is not directly applicable as we do not have access to the image-label annotations.
To address this problem, only one detection $d_{best}$, which has the highest probability $p$ among all detections, can be pseudo-labeled.
The results are shown as \texttt{PL w/o label} and \texttt{DT+PL w/o label} in \Tref{tbl:ap_on_each_class}.
Fine-tuning the FSD on the images labeled by this method harms the performance as the result of pseudo-labeling contains a lot of inaccuracy.
Therefore, image-level annotations in the target domain are essential for PL, which greatly improves the detection performance.

\paragraph{Generality across Detectors}
\begin{table}[t]
  \caption{Results of our methods on the different baseline FSDs in terms of mAP [\%] in \datasetclipart.}
  \label{tbl:comparison_of_detectors}
  \centering
  \begin{tabular}{@{}lccc@{}} \toprule
  	Method & SSD300 & YOLOv2 & Faster R-CNN \\ \midrule
	Baseline & 26.8 & 25.5 & 26.2\\
	DT & 38.0 & 31.5 & 32.1 \\
	PL & 36.4 & 34.0 & 29.8 \\
	DT+PL & \textbf{46.0} & \textbf{39.9} & \textbf{34.9} \\ \midrule
	Ideal case & 55.4 & 51.2 & 50.0\\ \bottomrule
  \end{tabular}
\end{table}
We investigated our framework on other FSDs such as Faster R-CNN~\cite{ren2015faster} and YOLOv2~\cite{redmon2016yolo9000}.
Please refer to the supplementary material about details of the hyper-paramters.
The result further emphasizes the generality of our framework across all baseline FSDs, as shown in \Tref{tbl:comparison_of_detectors}.
We additionally found that the ensembling of SSD300 and Faster R-CNN yields 30.2~\% in terms of mAP and that of all the three FSDs yields 31.0~\% in terms of mAP, which is not so remarkable compared to the improvement by DT+PL.
The performance gain is significant in SSD300 compared to YOLOv2 and Faster R-CNN.
This result suggests the importance of data augmentation (e.g., the zoom-in and zoom-out features implemented in SSD300) during the process of training FSDs with pseudo-labeled annotations, which are often noisy and incomplete.
\begin{figure}[t]
{\captionsetup[subfigure]{justification=centering}
	\begin{minipage}[t]{0.49\linewidth}
		\centering
  		\includegraphics[width=1.1\hsize]{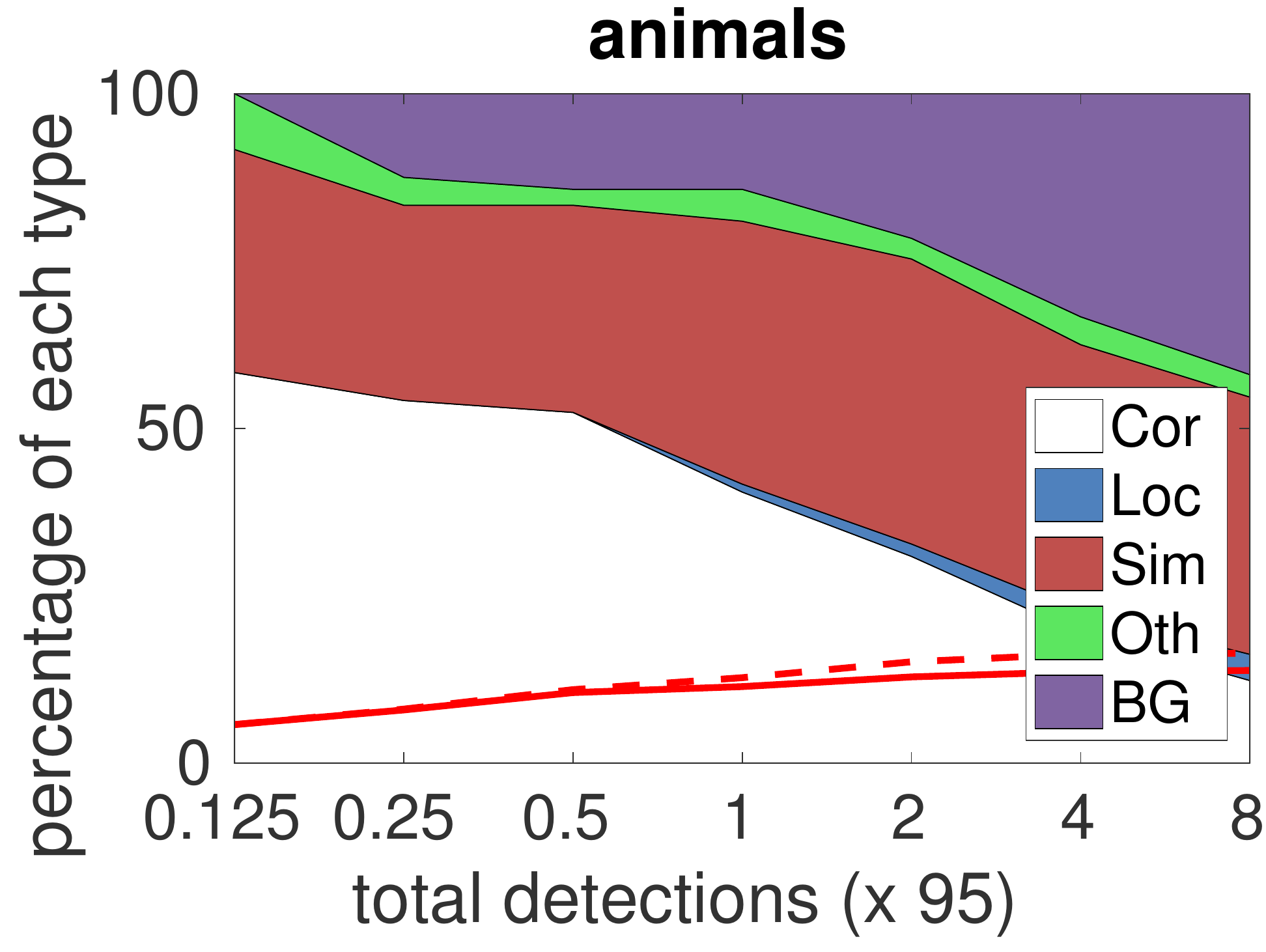}
	\end{minipage}
	\begin{minipage}[t]{0.49\linewidth}
		\includegraphics[width=1.1\hsize]{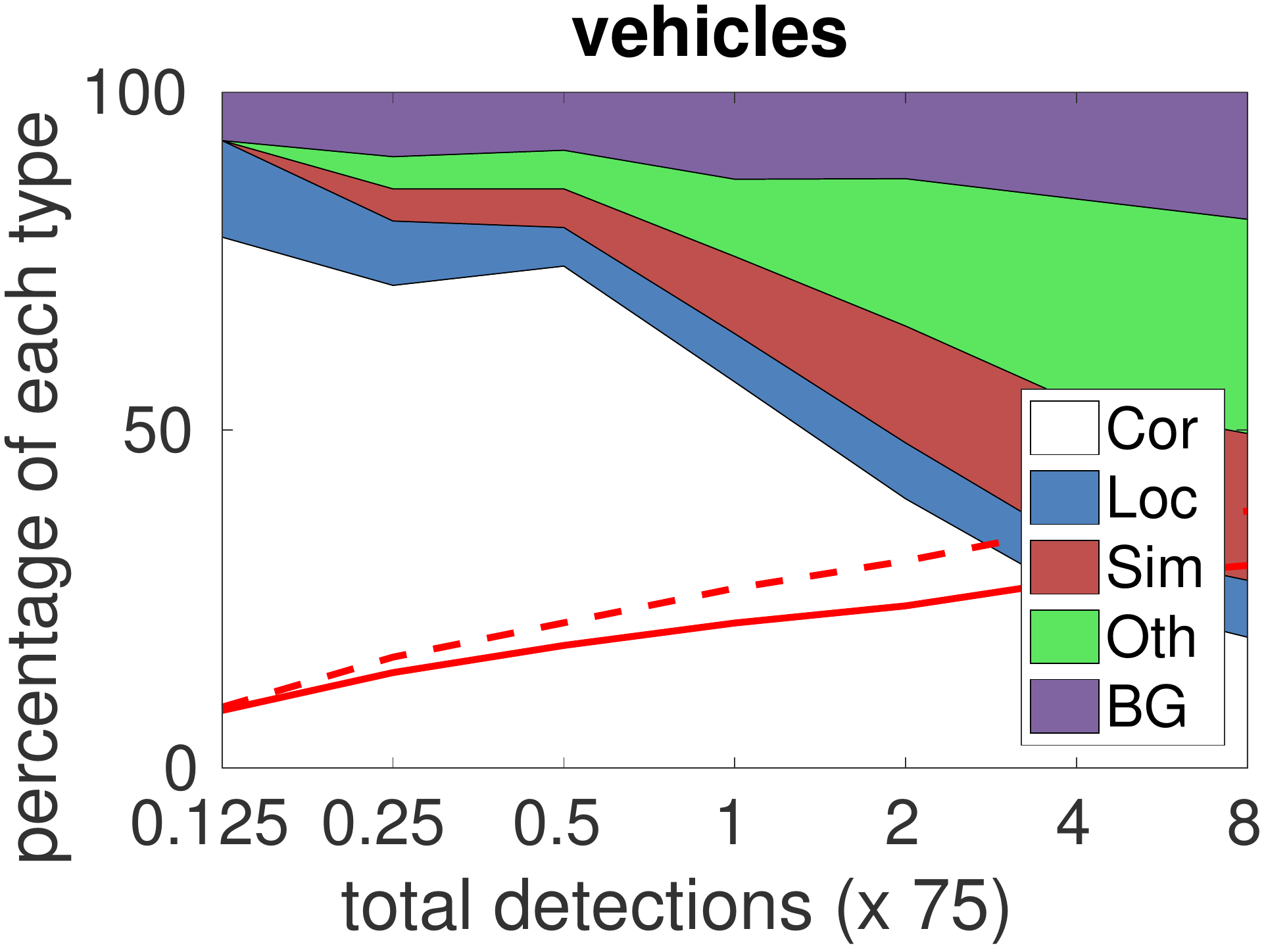}
	\end{minipage}
	\subcaption{Baseline}
	\begin{minipage}[t]{0.49\linewidth}
  		\centering
  		\includegraphics[width=1.1\hsize]{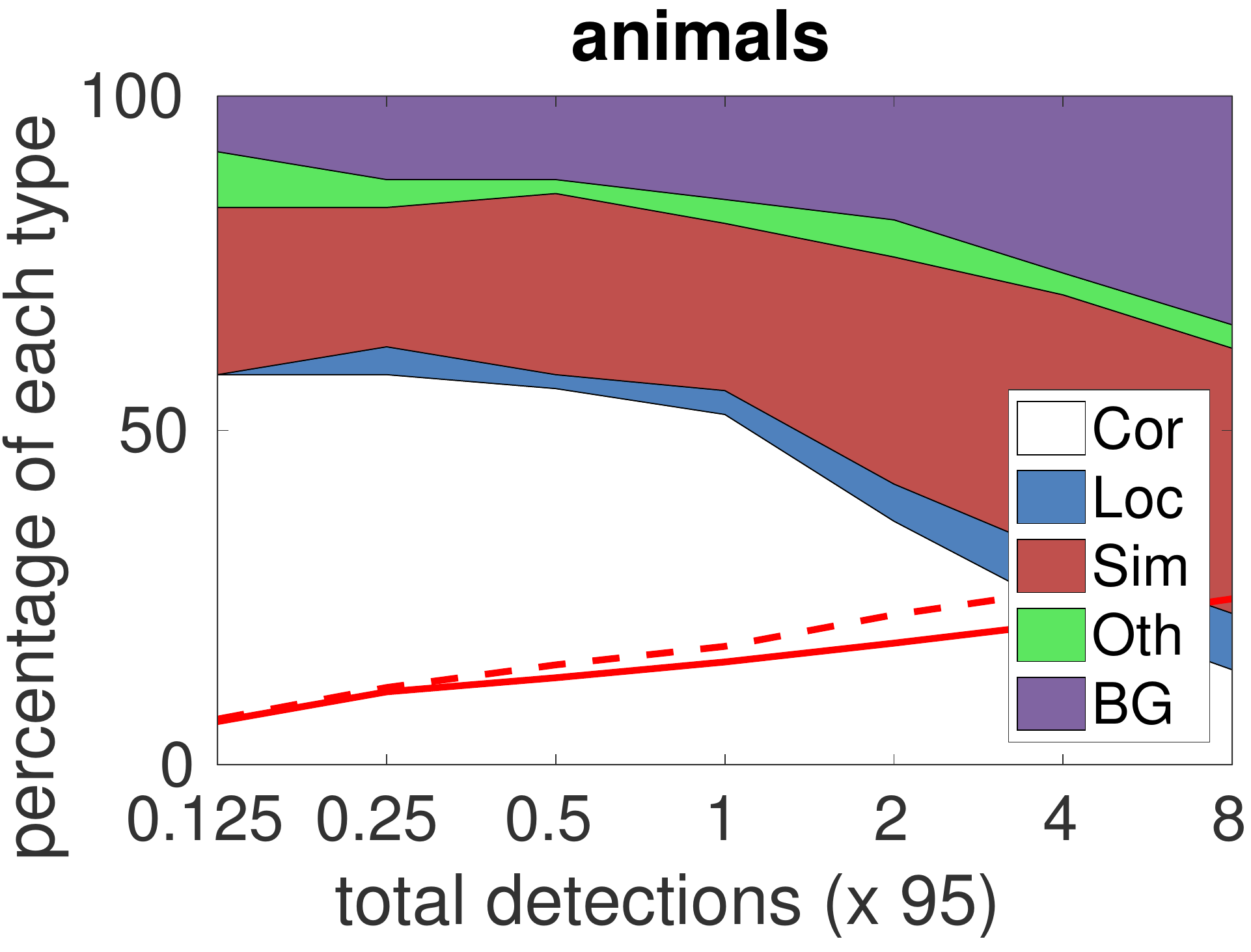}
  	\end{minipage}
	\begin{minipage}[t]{0.49\linewidth}
 		\includegraphics[width=1.1\hsize]{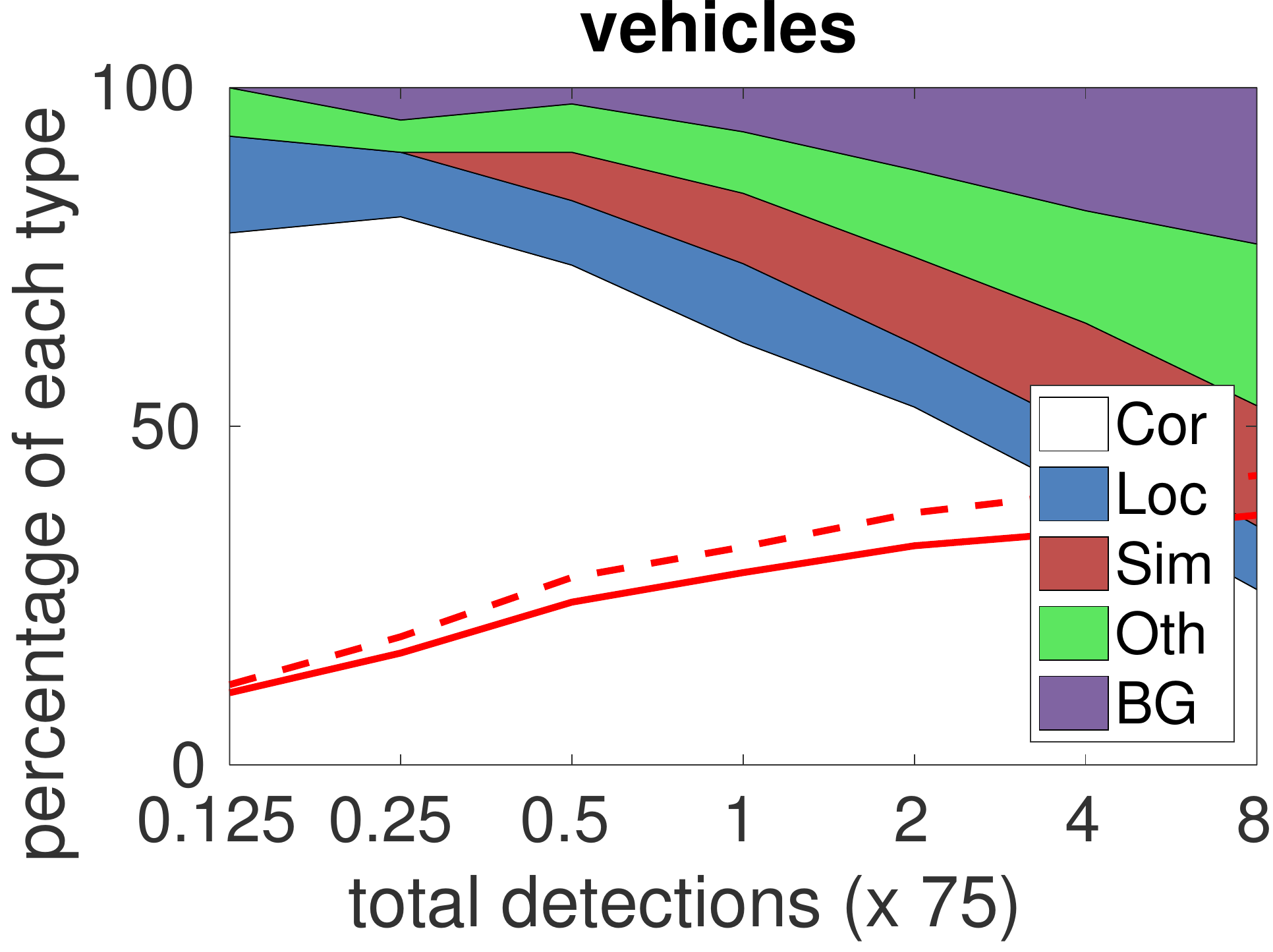}
 	\end{minipage}
    \subcaption{DT}
	\begin{minipage}[t]{0.49\linewidth}
		\centering
  		\includegraphics[width=1.1\hsize]{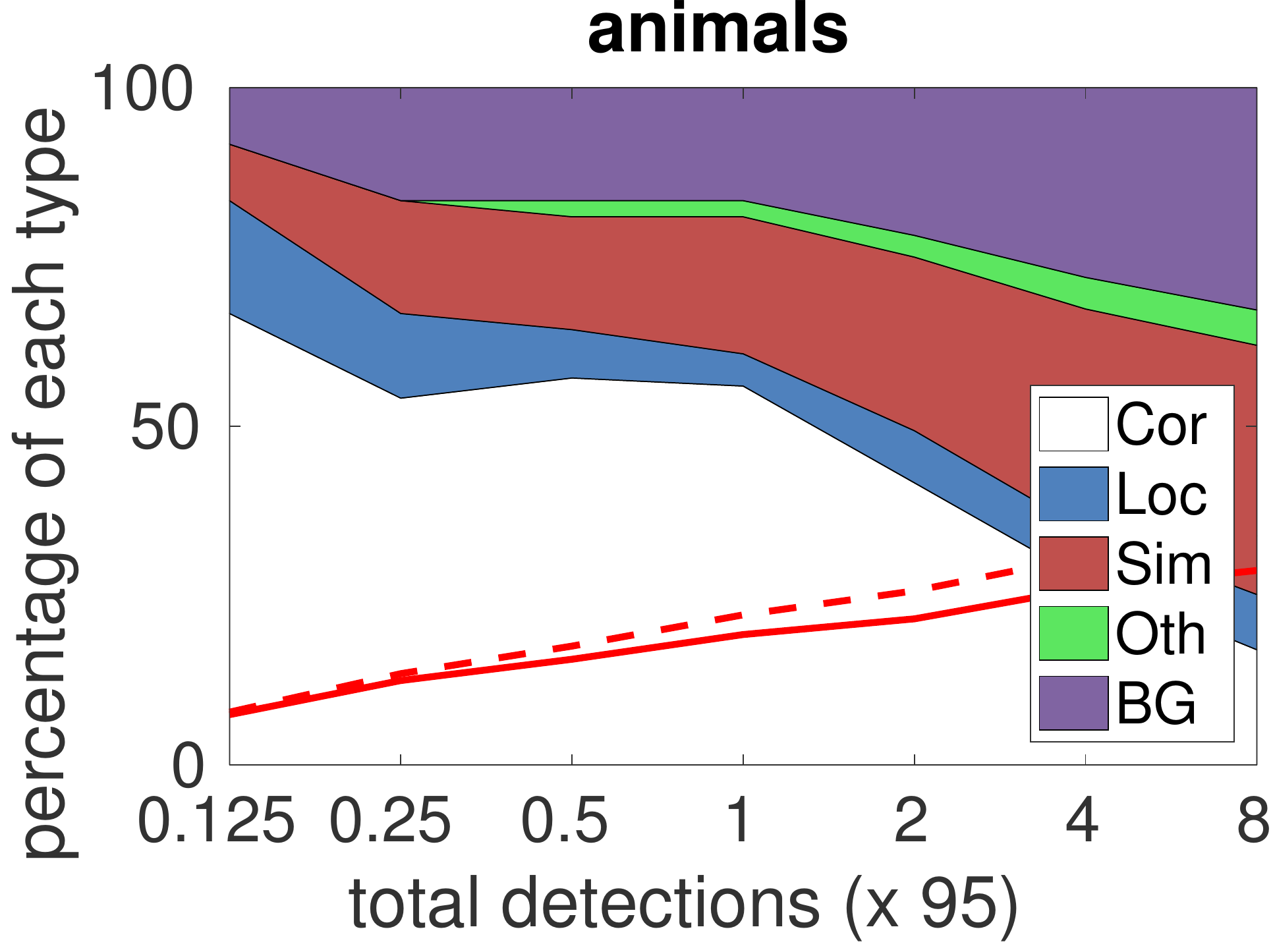}
  	\end{minipage}
  	\begin{minipage}[t]{0.49\linewidth}
  		\centering
  		\includegraphics[width=1.1\hsize]{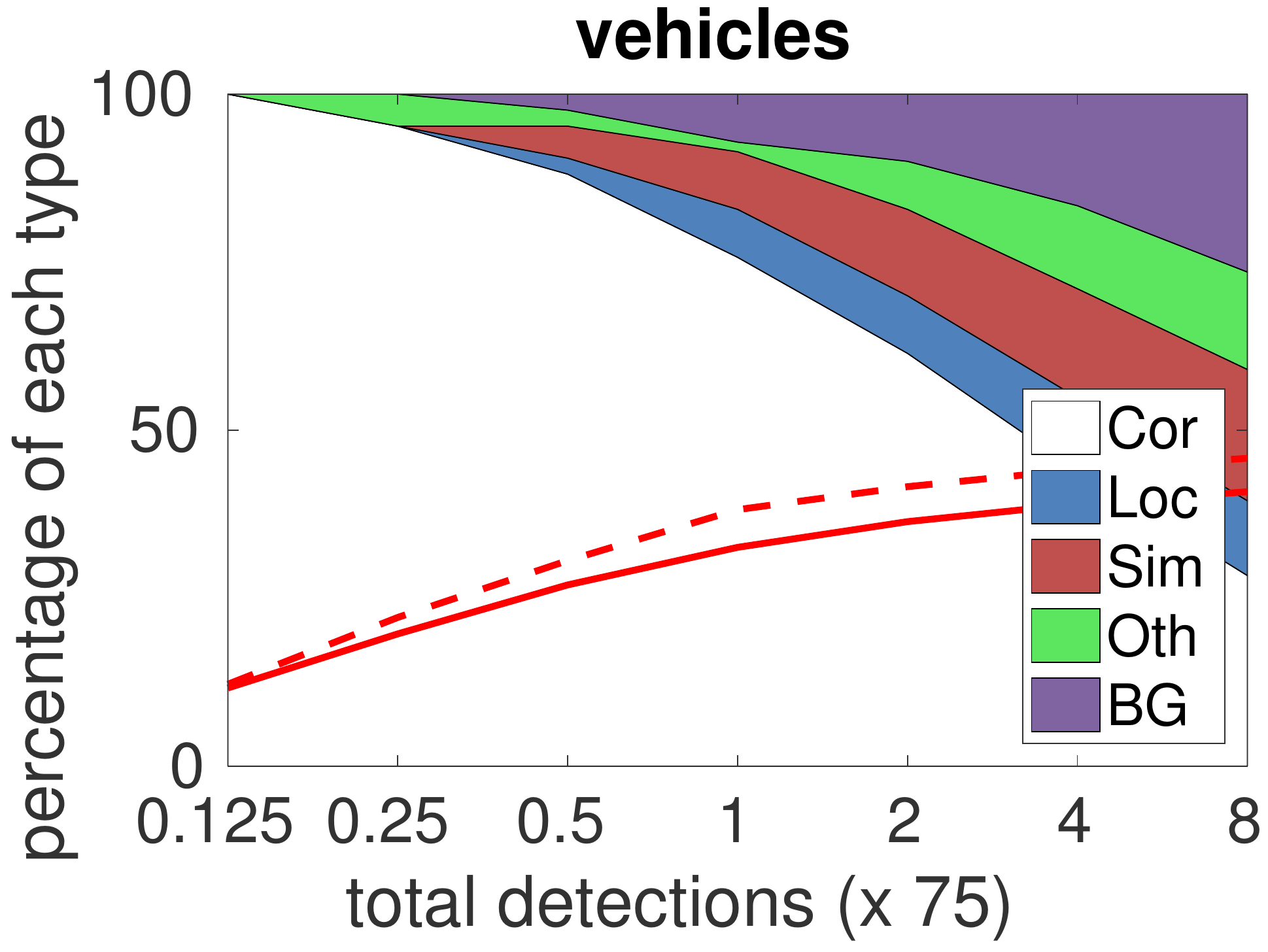}
  	\end{minipage}
    \subcaption{DT+PL}
}
\caption{Visualization of performance for various methods on animals and vehicles in the test set of \datasetclipart~using SSD300 as the baseline FSD.
 The solid red line and dashed red line reflect the change of recall with strong criteria (0.5 jaccard overlap) and weak criteria (0.1 jaccard overlap) as the number of detections increases, respectively.
}
\label{fig:visualization_of_performance}
\end{figure}

\begin{table}[t]
    \caption{Comparison in terms of AP [\%] using SSD300 as the baseline FSD in \datasetwatercolor.}
    \label{tbl:ap_on_each_class_bam_watercolor}
        \tabcolsep=1.5pt
        \centering
        \begin{tabular}{@{}lccccccc@{}}\toprule
        	& \multicolumn{6}{c}{AP for each class} & \\ \cmidrule(r){2-7}
            Method & bike & bird & car & cat & dog & person & mAP \\ \midrule
            Baseline & 79.8 & 49.5 & 38.1 & 35.1 & 30.4 & 65.1 & 49.6 \\ \midrule
            \textit{Compared} & & & & & & & \\
            \lvl WSDDN~\cite{bilen2016weakly} & 1.5 & 26.0 & 14.6 & 0.4 & 0.5 & 33.3 & 12.7 \\
            \lvl CLNet~\cite{kantorov2016contextlocnet} & 4.5 & 27.9 & 19.6 & 14.3 & 6.4 & 31.4 & 17.4 \\
            \lvl Ensemble & 79.8 & 49.6 & 38.1 & 35.2 & 30.4 & 58.7 & 48.6\\
            \lvl ADDA~\cite{tzeng2017adversarial} & 79.9 & 49.5 & 39.5 & 35.3 & 29.4 & 65.1 & 49.8 \\ \midrule
            \textit{Proposed} & & & & & & & \\
            \lvl PL & 76.3 & \textbf{54.9} & \textbf{46.6} & \textbf{37.5} & 36.9 & 71.7 & 54.0 \\
            \lvl DT & \textbf{82.8} & 47.0 & 40.2 & 34.6 & 35.3 & 62.5 & 50.4 \\
            \lvl DT+PL & 76.5 & \textbf{54.9} & 46.0 & 37.4 & \textbf{38.5} & \textbf{72.3} & \textbf{54.3}\\ \midrule
            \lvl PL~(+extra) & 84.8 & 57.7 & 48.0 & 44.9 & 46.6 & 72.6 & 59.1\\
            \lvl DT+PL~(+extra) & 86.3 & 57.3 & 48.5 & 43.0 & 46.5 & 73.2 & 59.1\\ \midrule
            Ideal case & 76.0 & 60.0 & 52.7 & 41.0 & 43.8 & 77.3 & 58.4 \\ \bottomrule
        \end{tabular}
\end{table}

\begin{table}[t]
    \caption{Comparison in terms of AP [\%] using SSD300 as the baseline FSD in \datasetcomic.}
    \label{tbl:ap_on_each_class_bam_comic}
        \tabcolsep=1.5pt
        \centering
        \begin{tabular}{@{}lccccccc@{}}\toprule
        	& \multicolumn{6}{c}{AP for each class} & \\ \cmidrule(r){2-7}
        	Method & bike & bird & car & cat & dog & person & mAP \\ \midrule
        	Baseline & 43.9 & 10.0 & 19.4 & 12.9 & 20.3 & 42.6 & 24.9 \\ \midrule
        	\textit{Compared} & & & & & & & \\
            \lvl WSDDN~\cite{bilen2016weakly} & 1.5 & 0.1 & 11.9 & 6.9 & 1.4 & 12.1 & 5.6 \\
            \lvl CLNet~\cite{kantorov2016contextlocnet} & 0.0 & 0.0 & 2.0 & 4.7 & 1.2 & 14.9 & 3.8 \\
            \lvl Ensemble & 44.0 & 10.0 & 19.4 & 14.5 & 20.7 & 42.9 & 25.3 \\
            \lvl ADDA~\cite{tzeng2017adversarial} & 39.5 & 9.8 & 17.2 & 12.7 & 20.4 & 43.3 & 23.8 \\ \midrule
            \textit{Proposed} & & & & & & &\\
            \lvl PL & 52.9 & 13.7 & 35.3 & 16.2 & 28.9 & 50.8 & 32.9 \\
            \lvl DT & 43.6 & 13.6 & 30.2 & 16.0 & 26.9 & 48.3 & 29.8\\
            \lvl DT+PL & \textbf{55.2} & \textbf{18.5} & \textbf{38.2} & \textbf{22.9} & \textbf{34.1} & \textbf{54.5} & \textbf{37.2} \\ \midrule
            \lvl PL~(+extra) & 53.4 & 19.0 & 35.0 & 30.0 & 30.5 & 53.7 & 36.9 \\
            \lvl DT+PL~(+extra) & 56.6 & 24.0 & 40.7 & 35.8 & 39.0 & 57.3 & 42.2 \\ \midrule
            Ideal case & 55.9 & 26.8 & 40.4 & 42.3 & 43.0 & 70.1 & 46.4 \\ \bottomrule
        \end{tabular}
\end{table}

\begin{figure*}[t]
	\begin{minipage}[t]{0.27\hsize}
		\centering
    	\includegraphics[height=5.5cm]{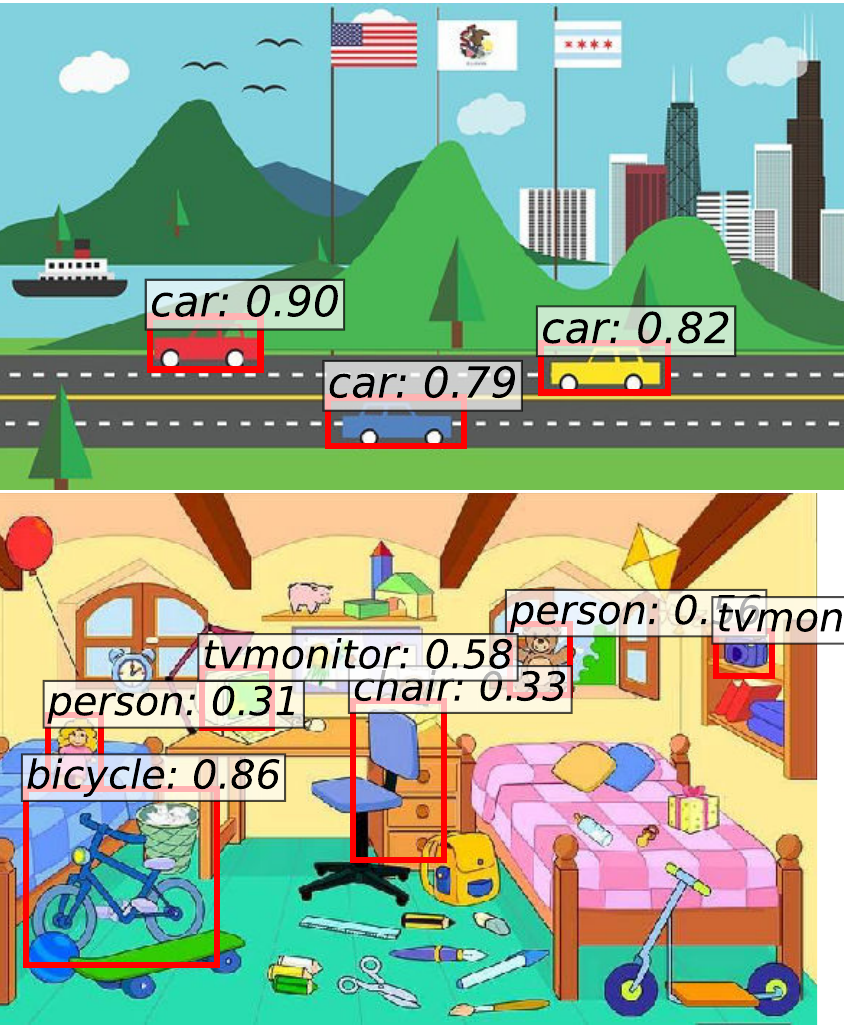}
		\subcaption{\datasetclipart}
		\label{fig:gt_clipart}
	\end{minipage}
	\hfill
	\begin{minipage}[t]{0.45\hsize}
		\centering
    	\includegraphics[height=5.5cm]{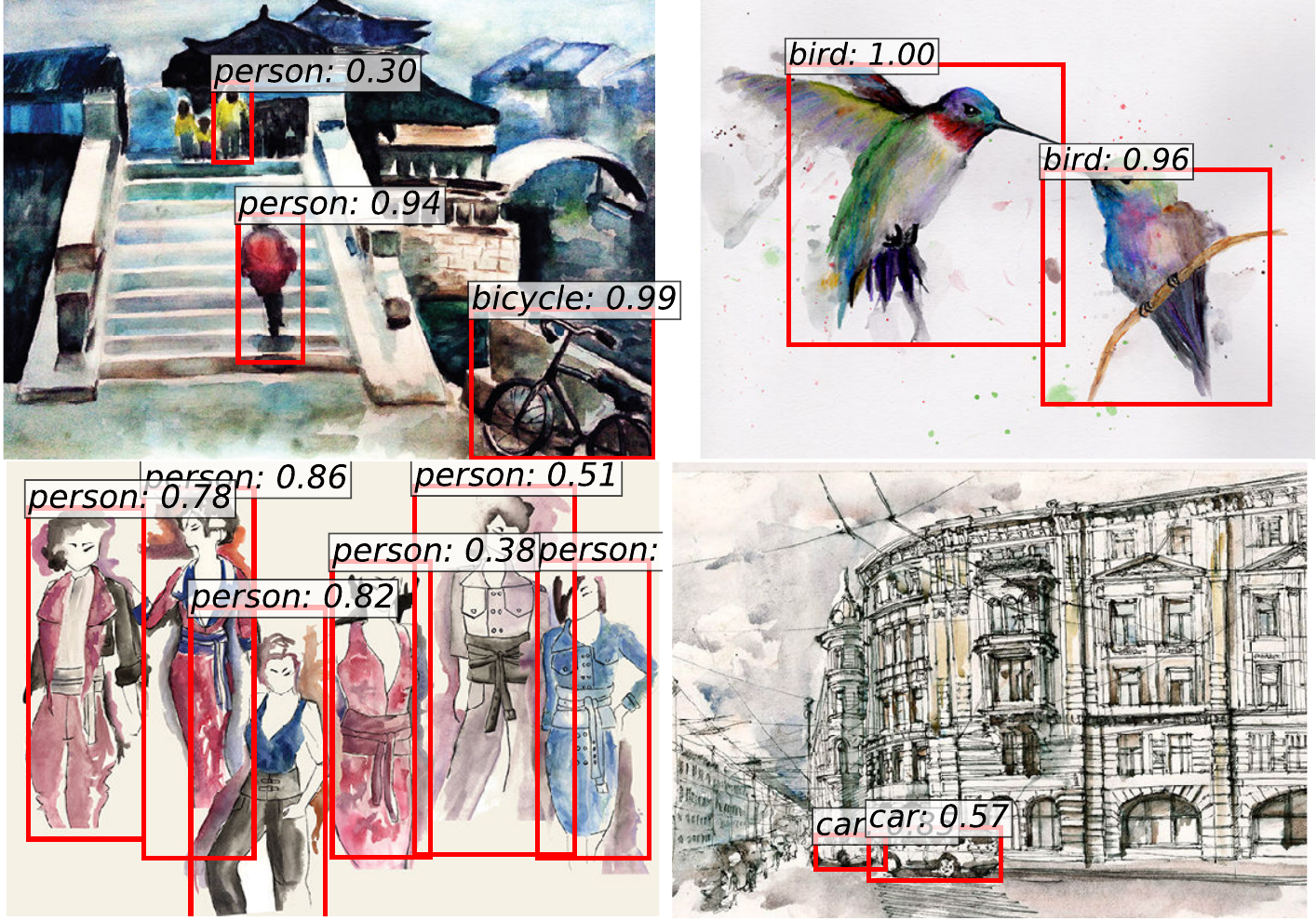}
    	\vspace{-0.4cm}
		\subcaption{\datasetwatercolor}
		\label{fig:gt_watercolor}
	\end{minipage}
	\hfill
	\begin{minipage}[t]{0.26\hsize}
		\centering
    	\includegraphics[height=5.5cm]{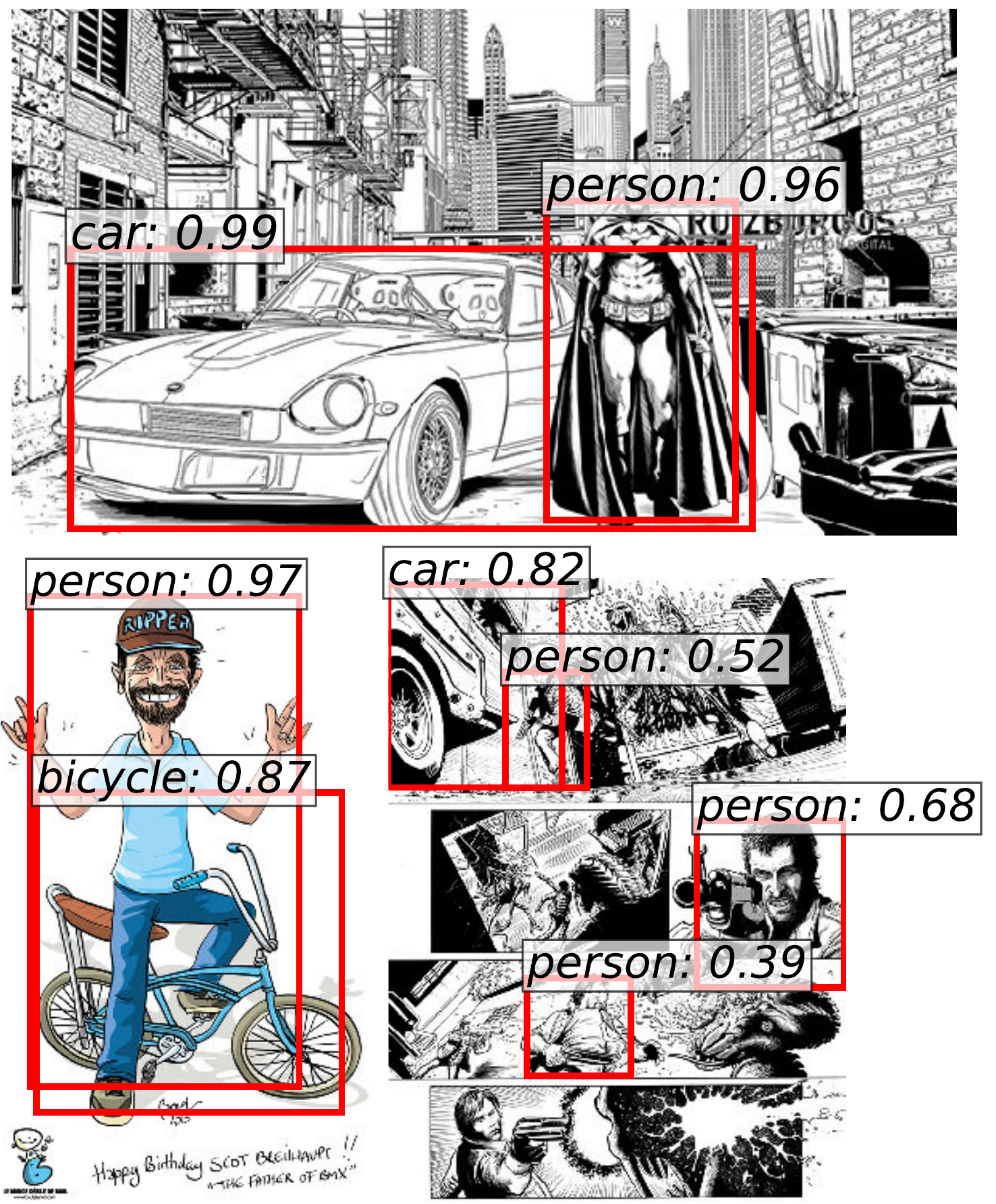}
    	\vspace{-0.4cm}
		\subcaption{\datasetcomic}
		\label{fig:gt_comic}
	\end{minipage}
    \caption{Example outputs for our DT+PA in the test set of each dataset. We only show windows whose scores are over 0.25 to maintain visibility.}
    \label{fig:results}
\end{figure*}
\begin{figure}[t]
	\centering
   	\includegraphics[width=\hsize]{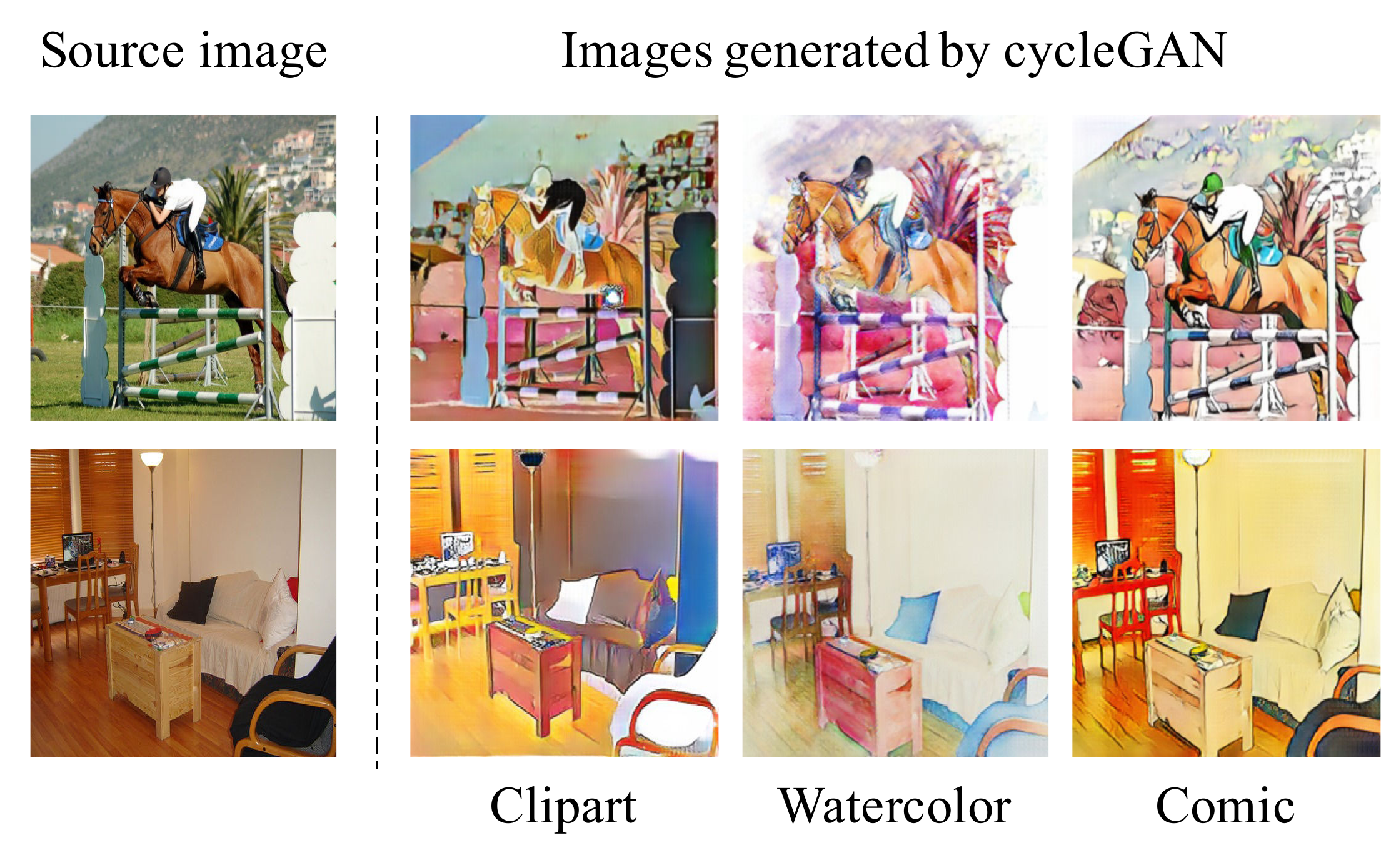}
    \caption{Example images generated by DT.}
    \label{fig:cycleGAN_results}
\end{figure}
\paragraph{Performance Analysis Focusing on Errors}
\label{subsec:performance_analysis}
The tool from~\cite{hoiem2012diagnosing} was used to understand the type of the detection error that is reduced by our methods.
The classes within the brackets were regarded as the same category: \{all vehicles\}, \{all animals including person\}, \{chair, dining table, sofa\}(furniture), \{aeroplane, bird\}(air objects).
Considering the class, the category, and the IoU between the predicted bounding box and the ground truth bounding box, the detections were classified into five groups as listed below:
\begin{itemize}[wide=0pt]
\setlength{\itemsep}{0cm}
    \item Correct (Cor): correct class and $IoU>.5$
    \item Localization (Loc): correct class, misaligned bounding box ($.1<IoU<.5$)
    \item Similar (Sim): wrong class, correct category, $IoU>.1$
    \item Other (Oth): wrong class, wrong category, $IoU>.1$
    \item Background (BG): $IoU<.1$ for any object
\end{itemize}
\Fref{fig:visualization_of_performance} shows the example of the error analysis in the \datasetclipart test set.
Comparing the baseline and DT, we observe that fine-tuning the FSD on images obtained by DT improves the detection performance, especially in less-confident detections.
Comparing DT and DT+PL, we observe that the confusion, which emerged with the other classes (Sim and Oth), especially in more confident detections, is greatly reduced by PL, which uses the image-level annotations in the target domain to remove such confusions with the FSD.

\subsection{Quantitative Results on \\ \datasetwatercolor~and~\datasetcomic}
\label{subsec:results_bam}
The comparison among our methods against the baseline FSD and the comparable methods is shown in \Tref{tbl:ap_on_each_class_bam_watercolor} and \Tref{tbl:ap_on_each_class_bam_comic}.
In \datasetwatercolor, the learning rate was set to $1.0 \times 10^{-6}$ as the fine-tuning overfitted in $1.0 \times 10^{-5}$ even in \texttt{Ideal case}.
Both our methods work in the two domains.

\texttt{+extra} in both tables indicates the use of extra ~\bam~images with raw noisy image-level labels of the target classes as described in \Sref{subsec:dataset_comic_and_watercolor}.
These images were pseudo-labeled and used for fine-tuning the FSD.
With a substantial number of images, the training of \texttt{+extra} methods underwent 30000 iterations.
The methods using extra noisy labels in~\bam~significantly improved the detection performance and sometimes proved to be better than \texttt{Ideal case} trained on 1,000 clean instance-level annotations.
Without any manual annotation, our framework can use large-scale images with noisy labels.

\subsection{Qualitative Results}
\label{subsec:qualitative_results}
\Fref{fig:cycleGAN_results} shows the example images generated by DT.
There was no mode collapse in the training of CycleGAN.
Visibly, the perfect mapping is not accomplished in this experiment as the representation gap between a natural image domain and the other domains used in this paper is too wide as compared with the gap between synthetic and real images tackled in recent studies, such as~\cite{shrivastava2017learning, bousmalis2017unsupervised}.
CycleGAN seems to transfer color and texture while keeping most of the edges and semantics of the input image.
The result of fine-tuning the FSD on these domain-transferred images in \Tref{tbl:ap_on_each_class}, \Tref{tbl:ap_on_each_class_bam_watercolor}, and \Tref{tbl:ap_on_each_class_bam_comic} confirms the validity of our domain transfer method.
Moreover, our methods are valid for various depiction styles as shown in \Fref{fig:results}.
For more results, please refer to the supplementary material.

\section{Discussion}
In PL, only the top-1 bounding box for each class is employed.
The other instances, if any, can be considered as negative samples.
This issue is our future work.
Moreover, if we could extract features with the same size corresponding to each detection, using the standard MIL paradigm in WSD such as~\cite{gokberk2014multi,li2016weakly}, we would improve the localization accuracy in pseudo-labeling, which is also our future work.

\section{Conclusion}
We proposed the novel task, cross-domain weakly supervised object detection, and the novel framework performing the two-step progressive domain adaptation to address this task.
To evaluate our methods, we constructed original datasets comprising images with instance-level annotations in three visual domains.
The results suggested that our methods were better than the other existing comparable methods and provided a simple but solid baseline.

\paragraph{Acknowledgments}
This work was partially supported by JST-CREST~(JPMJCR1686) and Microsoft IJARC~core13.
N. Inoue is supported by GCL program of The Univ. of Tokyo by JSPS.
R. Furuta is supported by the Grants-in-Aid for Scientific Research (16J07267) from JSPS.
{\small

}

\appendix
\section*{Appendix}
\section{Statistics of Our Datasets}
\begin{figure}[t]
\begin{minipage}[t]{\hsize}
  	  \centering
  	  \includegraphics[width=0.8\hsize]{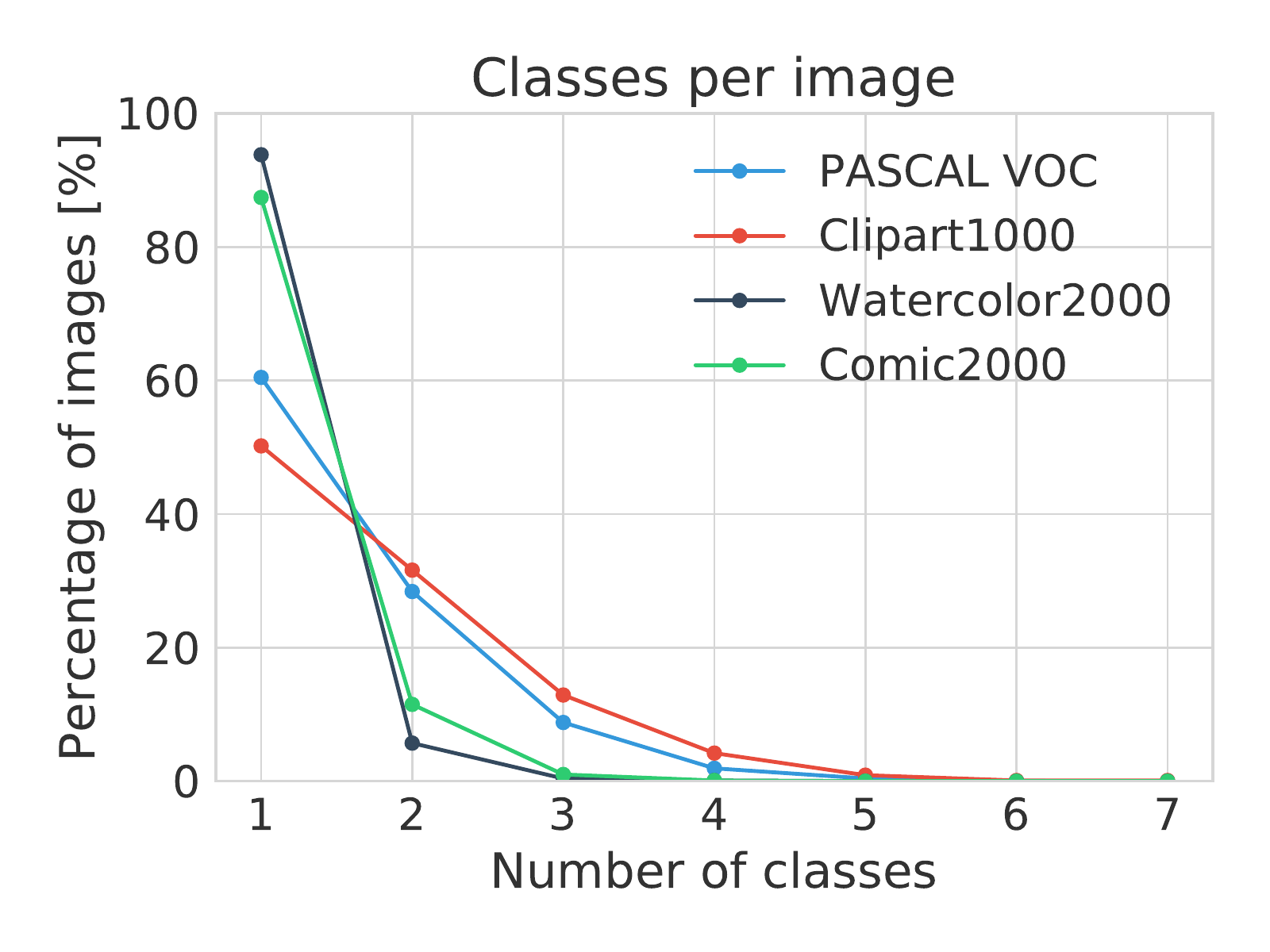}
  	  \subcaption{The number of classes per image in our datasets.}
\end{minipage}
\begin{minipage}[t]{\hsize}
	\centering
  	\includegraphics[width=0.8\hsize]{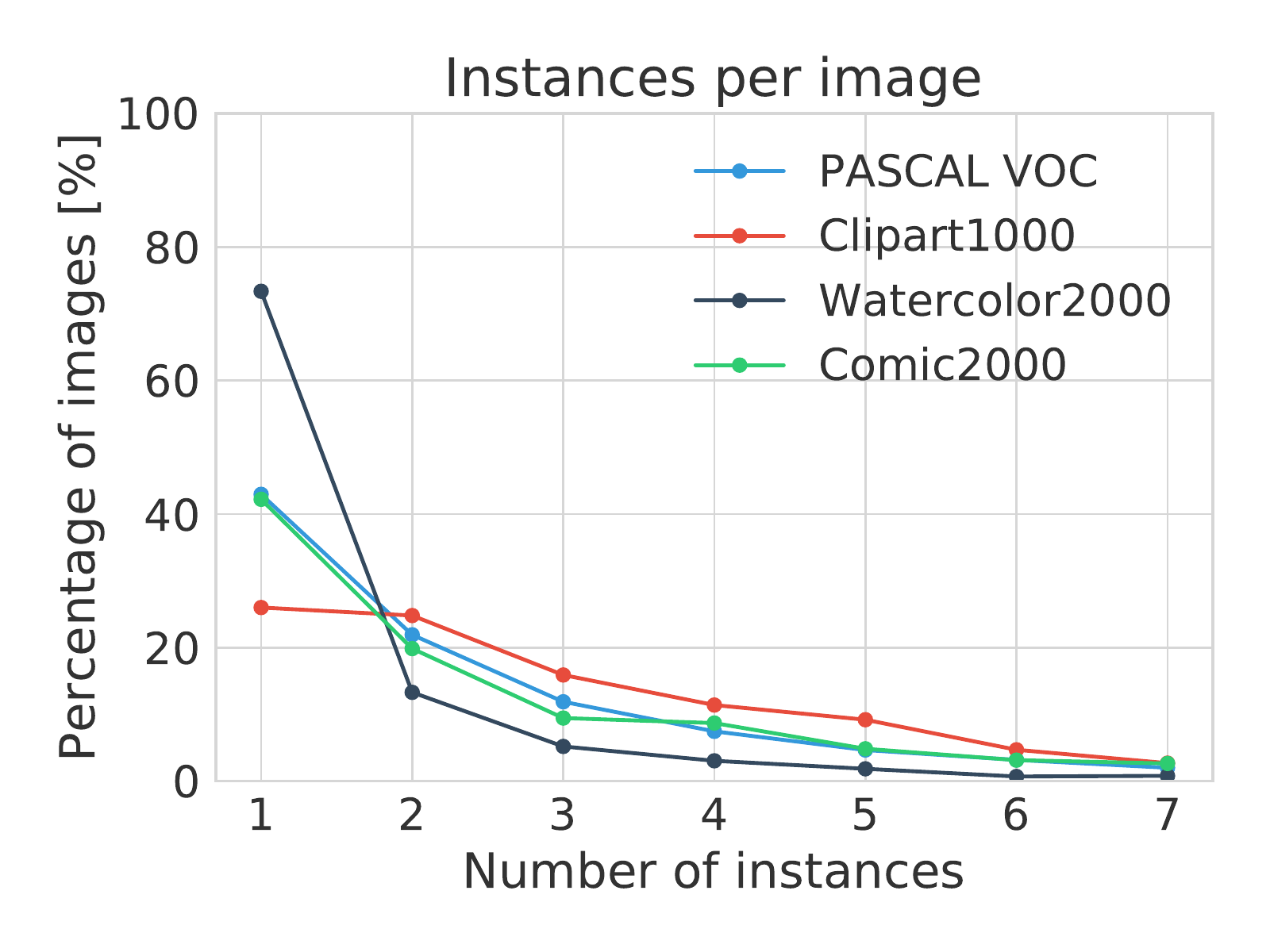}
  	\subcaption{The number of instances per image in our datasets.}
\end{minipage}
  \caption{Number of classes and instances in our datasets. For PASCAL~VOC, we used all the annotations including \texttt{difficult} boxes. Note that 
there are twenty object classes in PASCAL~VOC and \datasetclipart, and six object classes in \datasetwatercolor~and \datasetcomic.}
\label{fig:classes_and_instances_per_image}
\end{figure}
An important characteristic of our datasets is that they contain a sufficient number of objects.
The number of classes and instances per image is shown in \Fref{fig:classes_and_instances_per_image}.
For comparison, the figure contains the statistics of PASCAL~VOC~\cite{everingham2010pascal}, which is designed for detecting objects of twenty classes in natural images.
\datasetclipart~contains 1.7 classes and 3.2 instances per image.
\datasetclipart~contains almost the same number of classes and instances per image as PASCAL~VOC. 
The average number of classes and instances in \datasetclipart~is almost the same as that in PASCAL~VOC, which ensures the difficulty for the process of object detection.
\datasetwatercolor~contains 1.1 classes and 1.7 instances per image.
\datasetcomic~contains 1.1 classes and 3.2 instances per image.
Note that \datasetwatercolor~and~\datasetcomic~are for detecting the six classes.

As shown in \Tref{tbl:statistics_20class}, the distribution of the number of the instances for each class in \datasetclipart~is unbalanced, as is also seen in PASCAL~VOC~\cite{everingham2010pascal}.
In \Tref{tbl:statistics_6class}, the number of instances in \datasetwatercolor~and \datasetcomic~is shown.
In all datasets, the person class is dominant.

\begin{table}[t]
  \caption{The number of instances in \datasetclipart.}
  \label{tbl:statistics_20class}
  \centering
  \tabcolsep=2pt
  \begin{tabular}{@{}lc@{\hskip 0.5cm}lc@{}} \toprule
    Name & \#instances & Name & \#instances \\ \midrule
   	Aeroplane & 73 & Dining table & 115 \\
    Bicycle & 36 & Dog & 54 \\ 
    Bird & 265 & Horse & 79\\
    Boat & 129 & Motorbike & 17\\
    Bottle & 121 & Person & 1185\\
    Bus & 21 & Potted plant & 178\\
    Car & 202 & Sheep & 76\\
    Cat & 50 & Sofa & 52\\
    Chair & 340 & Train & 46\\
    Cow & 46 & TV/monitor & 80\\
    \bottomrule
  \end{tabular}
\end{table}

\begin{table}[t]
  \caption{The number of instances in \datasetwatercolor~and \datasetcomic.}
  \label{tbl:statistics_6class}
  \centering
  \tabcolsep=2pt
  \begin{tabular}{@{}lccccccc@{}} \toprule
    Dataset & Bicycle & Bird & Car & Cat & Dog & Person & Total \\ \midrule
    \datasetwatercolor & 27 & 486 & 101 & 102 & 116 & 2483 & 3315 \\
    \datasetcomic  & 87 & 270 & 107 & 233 & 192 & 5500 & 6389 \\ \bottomrule
  \end{tabular}
\end{table}

\begin{figure*}[t]
  \begin{minipage}[t]{0.265\hsize}
  	  \centering
  	  \includegraphics[width=\hsize]{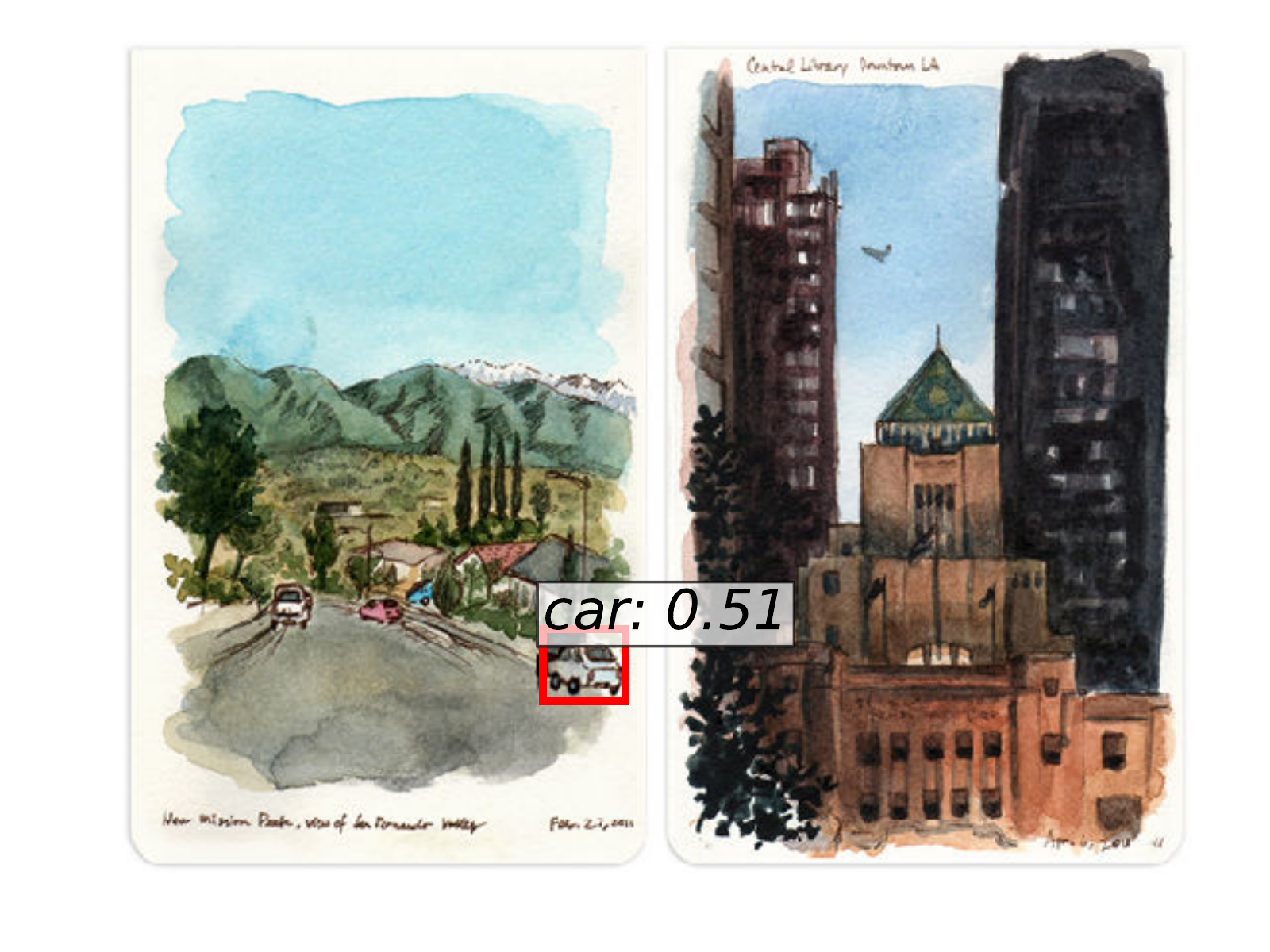}
  	  \subcaption{Ignoring small objects}
  	  \label{subfig:a}
  \end{minipage}
  \hfill
  \begin{minipage}[t]{0.265\hsize}
  	\includegraphics[height=4cm]{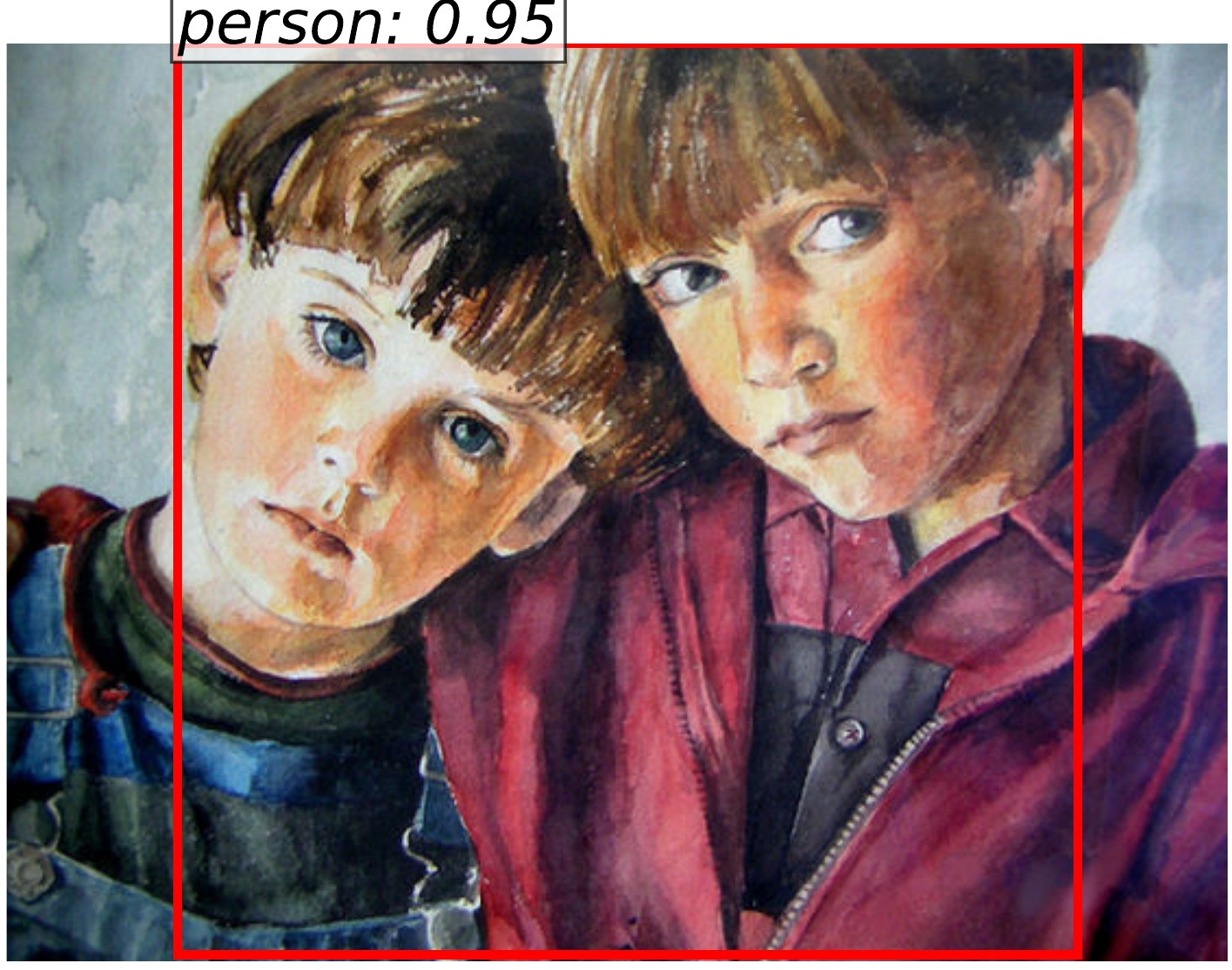}
  	\subcaption{Merging two objects}
  	\label{subfig:b}
  \end{minipage}
  \hfill
  \hfill
  \hfill
  \begin{minipage}[t]{0.175\hsize}
  	\includegraphics[width=\hsize]{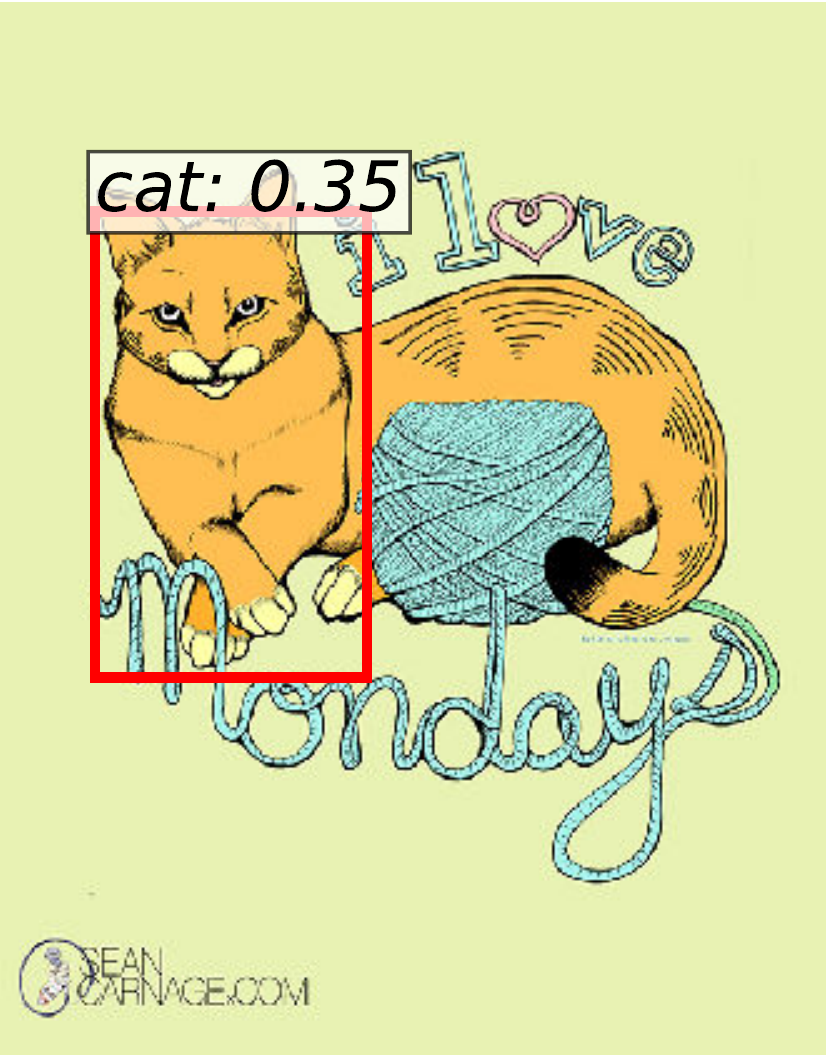}
  	\subcaption{Localizing most discriminative parts only}
  	\label{subfig:c}
  \end{minipage}
  \hfill
  \begin{minipage}[t]{0.225\hsize}
  	\includegraphics[width=\hsize]{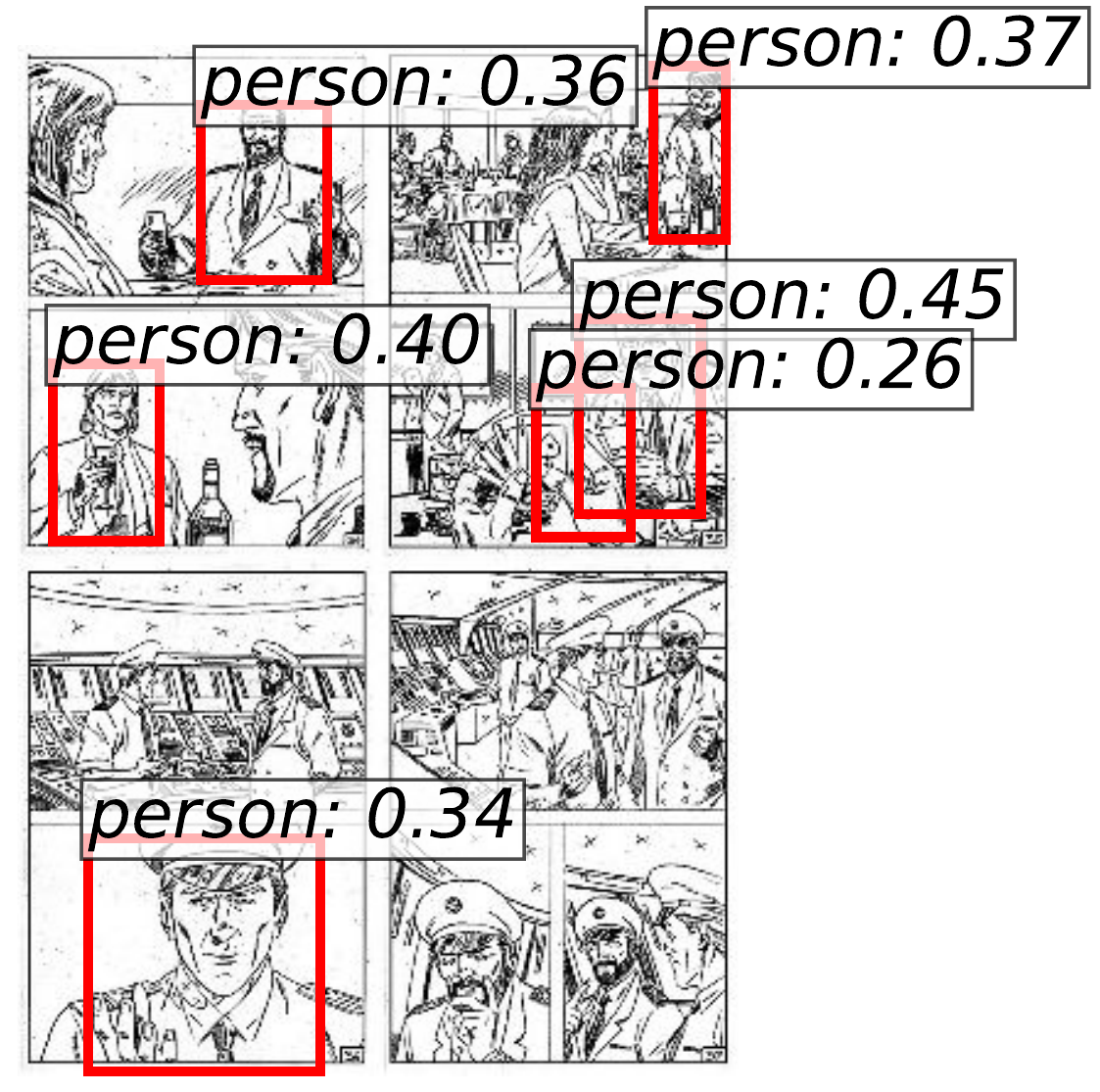}
  	\subcaption{Highly-deformed objects}
  	\label{subfig:d}
  \end{minipage}
  \caption{Typical detection errors by DT+PA using SSD300 as the baseline FSD. The images are from the test set of \datasetclipart~and \datasetcomic. }
\label{fig:visualization_of_typical_errors}
\end{figure*}

\section{Visualization of Detections}
\begin{figure*}[t]
  \centering
  \includegraphics[width=0.88\hsize]{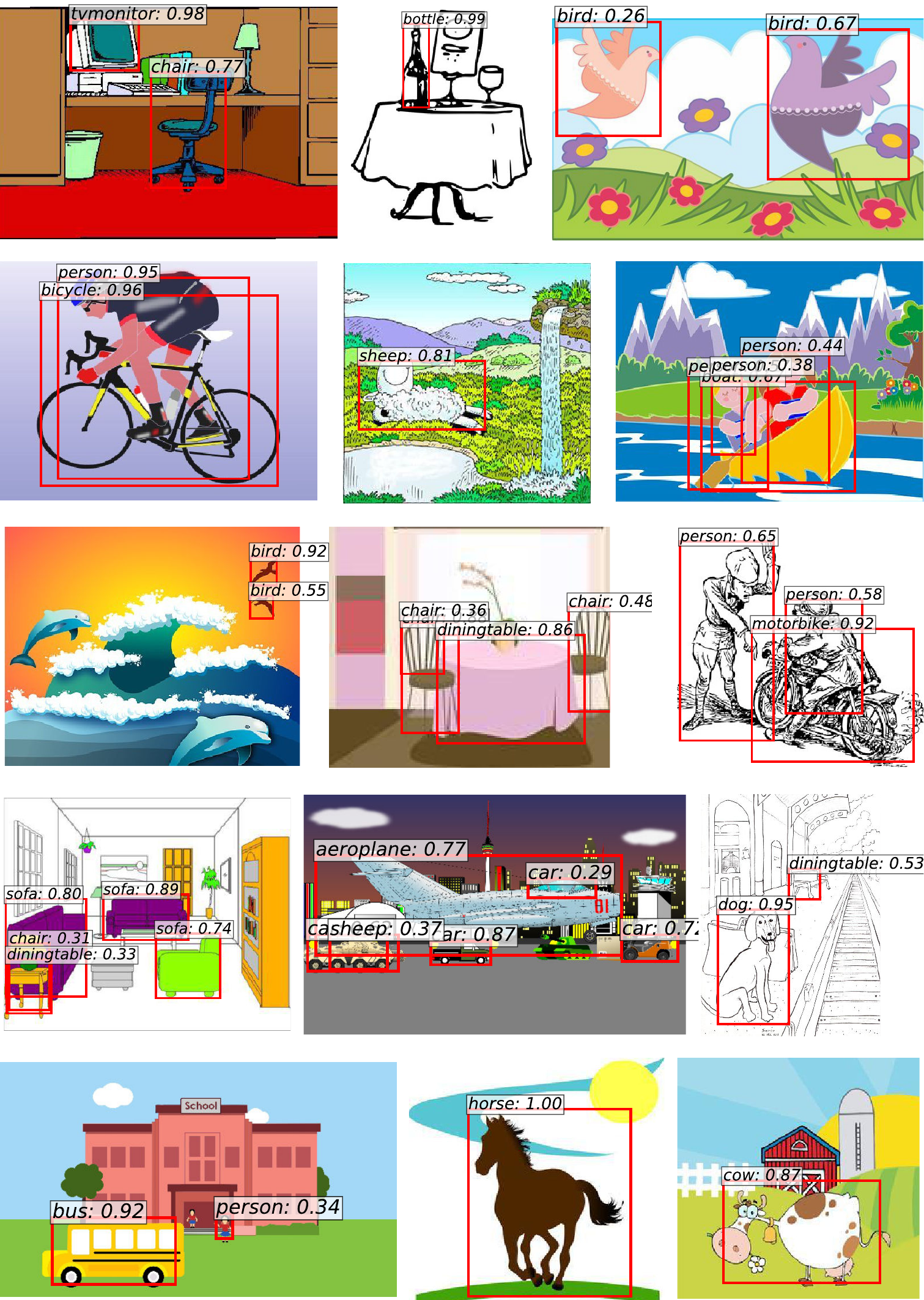}
  \caption{
Example outputs for our DT+PA using SSD300 as the baseline FSD in the test set of \datasetclipart.
We only showed windows whose scores are above 0.25 so as to maintain visibility.}
\label{fig:suppl_results_cliparts}
\end{figure*}

\begin{figure*}[t]
  \centering
  \includegraphics[width=0.9\hsize]{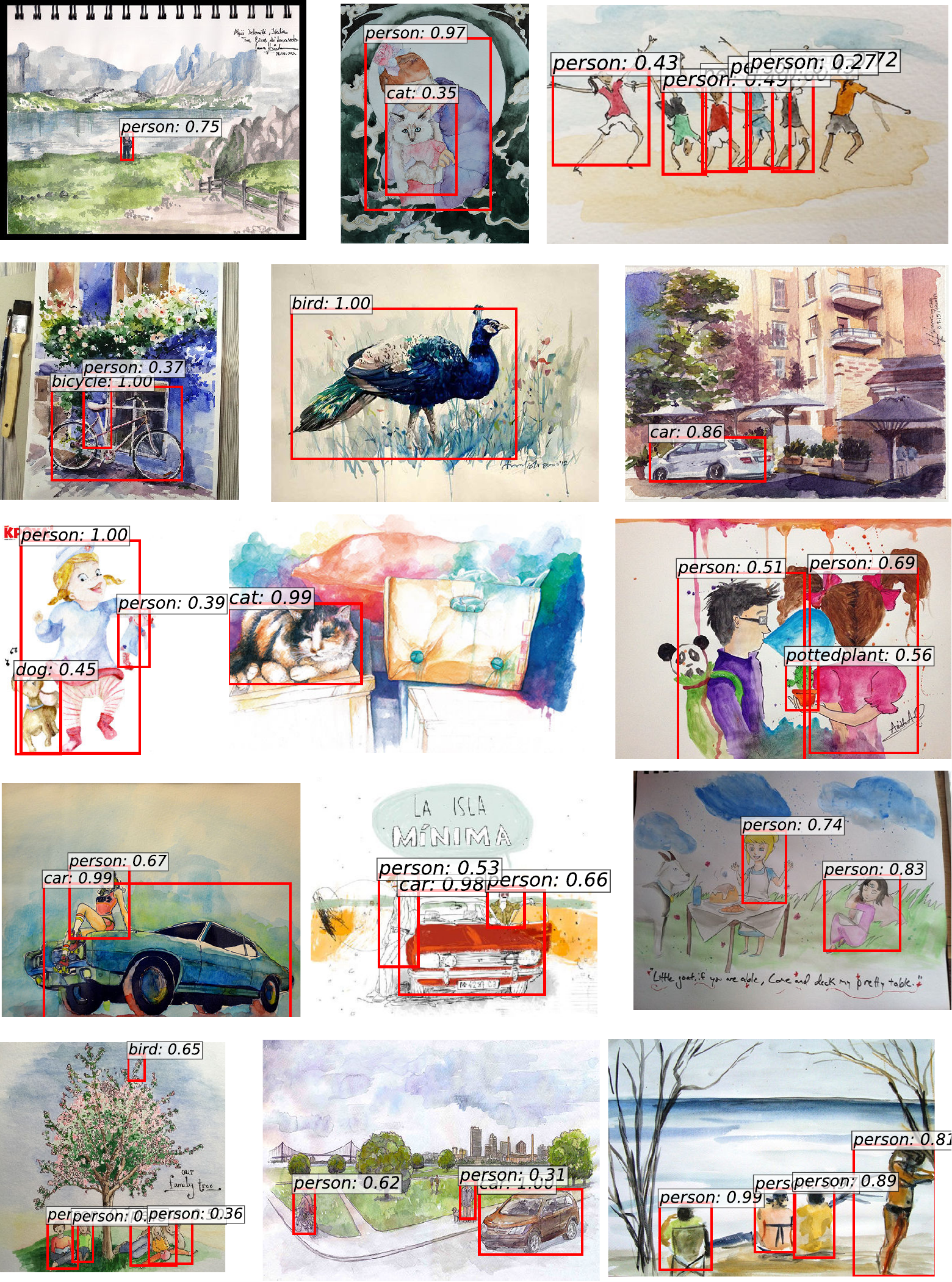}
  \caption{
Example outputs for our DT+PA using SSD300 as the baseline FSD in the test set of \datasetwatercolor.
We only showed windows whose scores are above 0.25 so as to maintain visibility.}
\label{fig:suppl_results_watercolor}
\end{figure*}

\begin{figure*}[t]
  \centering
  \includegraphics[width=0.9\hsize]{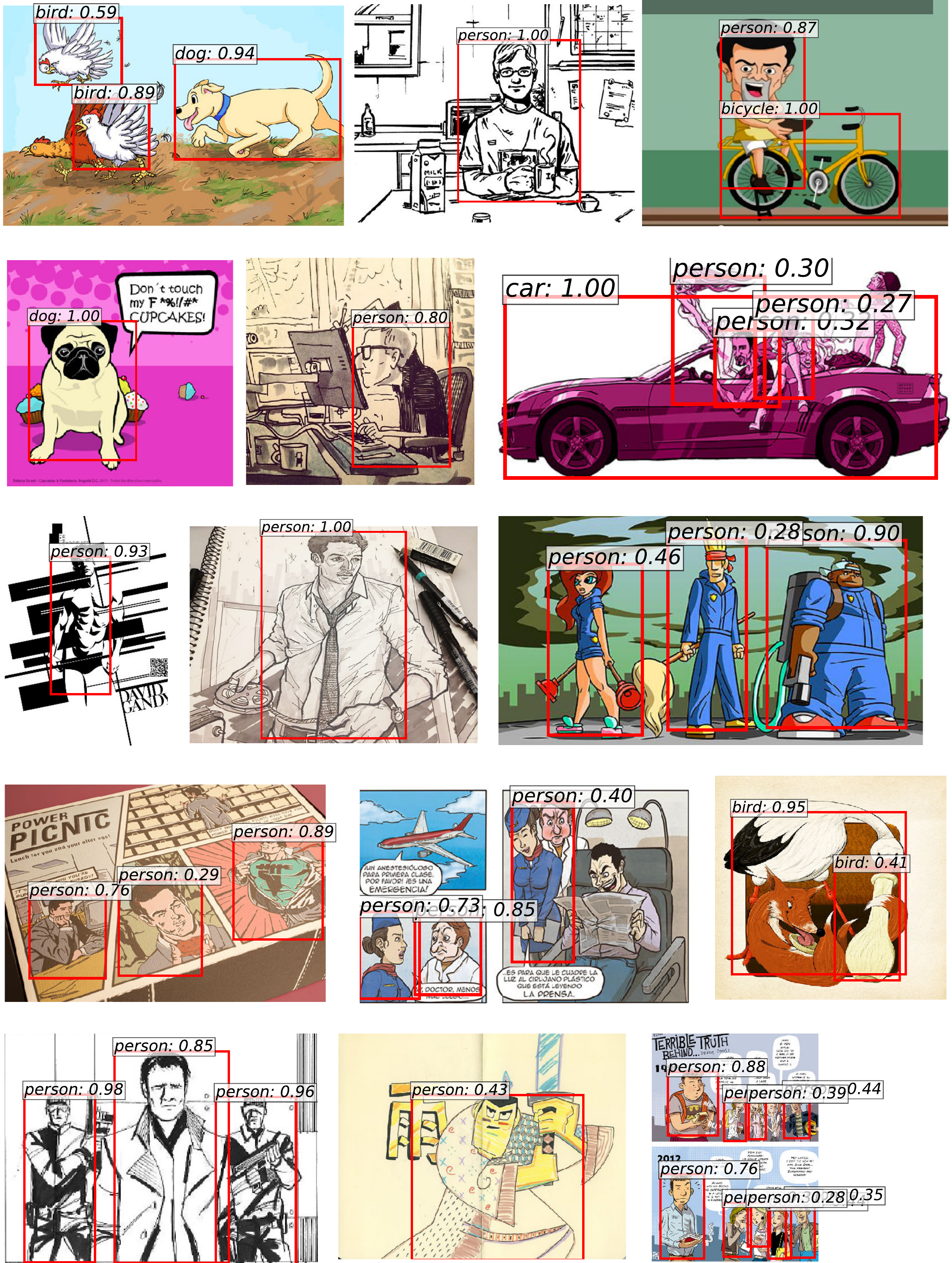}
  \caption{
Example outputs for our DT+PA using SSD300 as the baseline FSD in the test set of \datasetcomic.
We only showed windows whose scores are above 0.25 so as to maintain visibility.}
\label{fig:suppl_results_comic}
\end{figure*}

We discuss the detection results produced by our methods.
We will show the typical detection errors of our methods in \Fref{fig:visualization_of_typical_errors}.
The errors are often caused due to ignoring small objects (\Fref{subfig:a}), merging highly-overlapped objects which belong to the same object class (\Fref{subfig:b}), localizing only the most discriminative part of an object (\Fref{subfig:c}), or being unable to recognize highly-deformed objects (\Fref{subfig:d}).
The detections results obtained by our methods are shown in \Fref{fig:suppl_results_cliparts}, \Fref{fig:suppl_results_watercolor}, and \Fref{fig:suppl_results_comic}.
We confirm that our method is generally applicable and valid for various depiction styles.

\section{Implementation Details}

\subsection{Domain Transfer}
All the images were loaded and resized to 286 $\times$ 286.
In the fine-tuning phase, the images were randomly cropped to the size 256 $\times$ 256.
In the test phase, the images were loaded, transferred, and converted back to the original size.
We used all 16,551 images in VOC2007-trainval and VOC2012-trainval and obtained the domain-transferred images.

\subsection{Configurations for training FSDs}
For YOLOv2~\cite{redmon2016yolo9000}, we used the original implementation and employed a learning rate of $1.0 \times 10^{-5}$.
The input images were resized to 416 $\times$ 416.
With the IoU threshold (0.45) and the confidence threshold (0.001) employed, YOLOv2 was fine-tuned for five epochs and one hundred epochs for the DT and other experiments, respectively.

For Faster R-CNN~\cite{ren2015faster}, we used the reimplementation provided in ChainerCV~\cite{ChainerCV2017}.
We employed a learning rate of $1.0 \times 10^{-5}$.
The length of the shorter edge of the input image was scaled to 600.
After the scaling, if the length of the longer edge was longer than 1,000, the image was scaled so that the length of the longer edge came down to 1,000.
With the IoU threshold (0.3) and the confidence threshold (0.05) employed,
Faster R-CNN was fine-tuned for one epoch and one hundred epochs for the DT and other experiments, respectively.

\begin{thebibliography}{10}\itemsep=-1pt

\bibitem{bilen2015weakly}
H.~Bilen, M.~Pedersoli, and T.~Tuytelaars.
\newblock Weakly supervised object detection with convex clustering.
\newblock In {\em CVPR}, 2015.

\bibitem{bilen2016weakly}
H.~Bilen and A.~Vedaldi.
\newblock Weakly supervised deep detection networks.
\newblock In {\em CVPR}, 2016.

\bibitem{bousmalis2017unsupervised}
K.~Bousmalis, N.~Silberman, D.~Dohan, D.~Erhan, and D.~Krishnan.
\newblock Unsupervised pixel-level domain adaptation with generative
  adversarial networks.
\newblock In {\em CVPR}, 2017.

\bibitem{castrejon2016learning}
L.~Castrejon, Y.~Aytar, C.~Vondrick, H.~Pirsiavash, and A.~Torralba.
\newblock Learning aligned cross-modal representations from weakly aligned
  data.
\newblock In {\em CVPR}, 2016.

\bibitem{chen2015webly}
X.~Chen and A.~Gupta.
\newblock Webly supervised learning of convolutional networks.
\newblock In {\em ICCV}, 2015.

\bibitem{everingham2010pascal}
M.~Everingham, L.~Van~Gool, C.~K. Williams, J.~Winn, and A.~Zisserman.
\newblock The pascal visual object classes (voc) challenge.
\newblock {\em IJCV}, 88(2), 2010.

\bibitem{felzenszwalb2010object}
P.~F. Felzenszwalb, R.~B. Girshick, D.~McAllester, and D.~Ramanan.
\newblock Object detection with discriminatively trained part-based models.
\newblock {\em TPAMI}, 32(9), 2010.

\bibitem{ganin2016domain}
Y.~Ganin, E.~Ustinova, H.~Ajakan, P.~Germain, H.~Larochelle, F.~Laviolette,
  M.~Marchand, and V.~Lempitsky.
\newblock Domain-adversarial training of neural networks.
\newblock {\em JMLR}, 17(59), 2016.

\bibitem{girshick2015fast}
R.~Girshick.
\newblock Fast {R-CNN}.
\newblock In {\em ICCV}, 2015.

\bibitem{girshick2014rich}
R.~Girshick, J.~Donahue, T.~Darrell, and J.~Malik.
\newblock Rich feature hierarchies for accurate object detection and semantic
  segmentation.
\newblock In {\em CVPR}, 2014.

\bibitem{gokberk2014multi}
R.~Gokberk~Cinbis, J.~Verbeek, and C.~Schmid.
\newblock Multi-fold {MIL} training for weakly supervised object localization.
\newblock In {\em CVPR}, 2014.

\bibitem{gretton2012kernel}
A.~Gretton, K.~M. Borgwardt, M.~J. Rasch, B.~Sch{\"o}lkopf, and A.~Smola.
\newblock A kernel two-sample test.
\newblock {\em JMLR}, 13(Mar), 2012.

\bibitem{hoffman2015detector}
J.~Hoffman, D.~Pathak, T.~Darrell, and K.~Saenko.
\newblock Detector discovery in the wild: Joint multiple instance and
  representation learning.
\newblock In {\em CVPR}, 2015.

\bibitem{hoiem2012diagnosing}
D.~Hoiem, Y.~Chodpathumwan, and Q.~Dai.
\newblock Diagnosing error in object detectors.
\newblock In {\em ECCV}, 2012.

\bibitem{kantorov2016contextlocnet}
V.~Kantorov, M.~Oquab, M.~Cho, and I.~Laptev.
\newblock {ContextLocNet}: Context-aware deep network models for weakly
  supervised localization.
\newblock In {\em ECCV}, 2016.

\bibitem{kingma2015adam}
D.~P. Kingma and J.~Ba.
\newblock Adam: A method for stochastic optimization.
\newblock In {\em ICLR}, 2015.

\bibitem{openimages}
I.~Krasin, T.~Duerig, N.~Alldrin, V.~Ferrari, S.~Abu-El-Haija, A.~Kuznetsova,
  H.~Rom, J.~Uijlings, S.~Popov, A.~Veit, S.~Belongie, V.~Gomes, A.~Gupta,
  C.~Sun, G.~Chechik, D.~Cai, Z.~Feng, D.~Narayanan, and K.~Murphy.
\newblock Openimages: A public dataset for large-scale multi-label and
  multi-class image classification.
\newblock {\em https://github.com/openimages}, 2017.

\bibitem{li2016weakly}
D.~Li, J.-B. Huang, Y.~Li, S.~Wang, and M.-H. Yang.
\newblock Weakly supervised object localization with progressive domain
  adaptation.
\newblock In {\em CVPR}, 2016.

\bibitem{li2016r}
Y.~Li, K.~He, J.~Sun, et~al.
\newblock R-{FCN}: Object detection via region-based fully convolutional
  networks.
\newblock In {\em NIPS}, 2016.

\bibitem{lin2017focal}
T.-Y. Lin, P.~Goyal, R.~Girshick, K.~He, and P.~Doll{\'a}r.
\newblock Focal loss for dense object detection.
\newblock In {\em ICCV}, 2017.

\bibitem{lin2014microsoft}
T.-Y. Lin, M.~Maire, S.~Belongie, J.~Hays, P.~Perona, D.~Ramanan,
  P.~Doll{\'a}r, and C.~L. Zitnick.
\newblock Microsoft coco: Common objects in context.
\newblock In {\em ECCV}, 2014.

\bibitem{liu2016ssd}
W.~Liu, D.~Anguelov, D.~Erhan, C.~Szegedy, S.~Reed, C.-Y. Fu, and A.~C. Berg.
\newblock {SSD}: Single shot multibox detector.
\newblock In {\em ECCV}, 2016.

\bibitem{long2015learning}
M.~Long, Y.~Cao, J.~Wang, and M.~I. Jordan.
\newblock Learning transferable features with deep adaptation networks.
\newblock In {\em ICML}, 2015.

\bibitem{long2016unsupervised}
M.~Long, H.~Zhu, J.~Wang, and M.~I. Jordan.
\newblock Unsupervised domain adaptation with residual transfer networks.
\newblock In {\em NIPS}, 2016.

\bibitem{ChainerCV2017}
Y.~Niitani, T.~Ogawa, S.~Saito, and M.~Saito.
\newblock {ChainerCV}: a library for deep learning in computer vision.
\newblock In {\em ACM Multimedia}, 2017.

\bibitem{papadopoulos2017extreme}
D.~P. Papadopoulos, J.~R. Uijlings, F.~Keller, and V.~Ferrari.
\newblock Extreme clicking for efficient object annotation.
\newblock In {\em ICCV}, 2017.

\bibitem{redmon2016yolo9000}
J.~Redmon and A.~Farhadi.
\newblock {YOLO9000: Better, Faster, Stronger}.
\newblock In {\em CVPR}, 2017.

\bibitem{ren2015faster}
S.~Ren, K.~He, R.~Girshick, and J.~Sun.
\newblock Faster {R-CNN}: Towards real-time object detection with region
  proposal networks.
\newblock In {\em NIPS}, 2015.

\bibitem{shrivastava2017learning}
A.~Shrivastava, T.~Pfister, O.~Tuzel, J.~Susskind, W.~Wang, and R.~Webb.
\newblock Learning from simulated and unsupervised images through adversarial
  training.
\newblock In {\em CVPR}, 2017.

\bibitem{song2014learning}
H.~O. Song, R.~B. Girshick, S.~Jegelka, J.~Mairal, Z.~Harchaoui, T.~Darrell,
  et~al.
\newblock On learning to localize objects with minimal supervision.
\newblock In {\em ICML}, 2014.

\bibitem{song2014weakly}
H.~O. Song, Y.~J. Lee, S.~Jegelka, and T.~Darrell.
\newblock Weakly-supervised discovery of visual pattern configurations.
\newblock In {\em NIPS}, 2014.

\bibitem{su2012crowdsourcing}
H.~Su, J.~Deng, and L.~Fei-Fei.
\newblock Crowdsourcing annotations for visual object detection.
\newblock In {\em AAAI workshop}, 2012.

\bibitem{tang2017multiple}
P.~Tang, X.~Wang, X.~Bai, and W.~Liu.
\newblock Multiple instance detection network with online instance classifier
  refinement.
\newblock In {\em CVPR}, 2017.

\bibitem{tokui2015chainer}
S.~Tokui, K.~Oono, S.~Hido, and J.~Clayton.
\newblock Chainer: a next-generation open source framework for deep learning.
\newblock In {\em NIPS workshop}, 2015.

\bibitem{tzeng2017adversarial}
E.~Tzeng, J.~Hoffman, K.~Saenko, and T.~Darrell.
\newblock Adversarial discriminative domain adaptation.
\newblock In {\em CVPR}, 2017.

\bibitem{uijlings2013selective}
J.~R. Uijlings, K.~E. Van De~Sande, T.~Gevers, and A.~W. Smeulders.
\newblock Selective search for object recognition.
\newblock {\em IJCV}, 104(2):154--171, 2013.

\bibitem{westlake2016detecting}
N.~Westlake, H.~Cai, and P.~Hall.
\newblock Detecting people in artwork with cnns.
\newblock In {\em ECCV workshop}, 2016.

\bibitem{wilber2017bam}
M.~J. Wilber, C.~Fang, H.~Jin, A.~Hertzmann, J.~Collomosse, and S.~Belongie.
\newblock {BAM!} the behance artistic media dataset for recognition beyond
  photography.
\newblock In {\em ICCV}, 2017.

\bibitem{wu2014learning}
Q.~Wu, H.~Cai, and P.~Hall.
\newblock Learning graphs to model visual objects across different depictive
  styles.
\newblock In {\em ECCV}, 2014.

\bibitem{zhu2017unpaired}
J.-Y. Zhu, T.~Park, P.~Isola, and A.~A. Efros.
\newblock Unpaired image-to-image translation using cycle-consistent
  adversarial networks.
\newblock In {\em ICCV}, 2017.

\end{thebibliography}
\end{document}